# Double Coupled Canonical Polyadic Decomposition for Joint Blind Source Separation

Xiao-Feng Gong, *Member, IEEE*, Qiu-Hua Lin, *Member, IEEE*, Feng-Yu Cong, *Senior Member, IEEE*, and Lieven De Lathauwer, *Fellow, IEEE*

*Abstract* — Joint blind source separation (J-BSS) is an emerging data-driven technique for multi-set data-fusion. In this paper, J-BSS is addressed from a tensorial perspective. We show how, by using second-order multi-set statistics in J-BSS, a specific double coupled canonical polyadic decomposition (DC-CPD) problem can be formulated. We propose an algebraic DC-CPD algorithm based on a coupled rank-1 detection mapping. This algorithm converts a possibly underdetermined DC-CPD to a set of overdetermined CPDs. The latter can be solved algebraically via a generalized eigenvalue decomposition based scheme. Therefore, this algorithm is deterministic and returns the exact solution in the noiseless case. In the noisy case, it can be used to effectively initialize optimization based DC-CPD algorithms. In addition, we obtain the deterministic and generic uniqueness conditions for DC-CPD, which are shown to be more relaxed than their CPD counterpart. We also introduce optimization based DC-CPD methods, including alternating least squares, and structured data fusion based methods. Experiment results are given to illustrate the superiority of DC-CPD over standard CPD based BSS methods and several existing J-BSS methods, with regards to uniqueness and accuracy.

## I. Introduction

CANONICAL polyadic decomposition (CPD) decomposes a higher-order tensor into a minimal number of rank-1 terms. Compared with its matrix counterpart, CPD is essentially unique under mild conditions, without the need to exploit prior knowledge or impose particular constraints, and this nice identifiability property makes it a fundamental tool that is of particular interest for blind source separation (BSS). CPD has indeed been successfully adopted in many BSS applications [1]–[8] including telecommunications, array signal processing, speech separation, and biomedical data analysis. Numerous CPD algorithms have been proposed in the literature [9]–[16], and its identifiability has been extensively studied, including its ability to deal with underdetermined BSS problems [6], [10]–[13], [17]–[22]. The incorporation of various constraints (e.g. nonnegativity, source independence, Vandermonde structure etc.) has also been investigated [23]–[28], aiming at improved performance in a variety of applications.

Recently, joint BSS (J-BSS) of multi-set signals has been considered in a number of applications such as multi-subject / multi-modal biomedical data fusion and BSS of transformed signals at multiple frequency bins for convolutive mixtures. These approaches usually assume dependence across datasets (inter-set dependence) and independence of latent sources within a dataset (intra-set independence), aiming at BSS of each individual dataset as well as indication of correspondences among decomposed components. A number of J-BSS methods have been proposed, e.g. joint and parallel independent component analysis (ICA) [29], independent vector analysis (IVA) [30]–[32], multi-set canonical correlation analysis (M-CCA) [33]–[35] and generalized joint diagonalization (GJD) [36]–[39]. We refer to [40] and references therein for an overview.

There are also tensor-based results that concern multi-set problems [41]–[59]. The concept of linked mode parallel factor analysis originated in [41]. Tensor probabilistic ICA (T-PICA) [42] has been extended and applied to multi-set fMRI via group T-PICA [43] and linked T-PICA [44], which assume shared source and shared mixing matrix, respectively. A simple case of coupled CPD (C-CPD) of two tensors has been discussed in [47]–[49] with applications to blind deconvolution of MIMO systems, joint EEG-MEG analysis, and array signal processing, respectively. Linked CPD with partially shared factors is addressed in [50]. A comprehensive study of uniqueness and an algebraic algorithm for the computation of C-CPD and coupled rank-$(L_{r,n}, L_{r,n}, 1)$ block term decomposition with one coupling factor matrix are provided in [51] and [52], respectively. Several applications of C-CPD in signal and array processing are discussed in [45], [46], [53]–[55]. Soft or flexible coupling has been considered as a way to deal with nonidentical but similar coupling factor matrices [56]–[58]. Recently, C-CPD with double coupling structure has been considered in [59].

However, the C-CPD approaches mentioned above are not specifically devised for J-BSS problems. More precisely, the

Manuscript received XXXX XX, XXXX. This research is funded by: (1) National natural science foundation of China (nos. 61671106, 61331019, 61379012, 81471742); (2) Scientific Research Fund of Liaoning Education Department (no. L2014016); (3) Fundamental Research Funds for the Central Universities (nos. DUT16QY07). (4) Research Council KU Leuven: C1 project C16/15/059-nD. (5) FWO: project G.0830.14N, G.0881.14N, EOS project G0F6718N (SeLMA); (6) EU: The research leading to these results has received funding from the European Research Council under the European Union's Seventh Framework Programme (FP7/2007-2013) / ERC Advanced Grant: BIOTENSORS (no. 339804). This paper reflects only the authors' views and the Union is not liable for any use that may be made of the contained information.

Xiao-Feng Gong and Qiu-Hua Lin are with the School of Information and Communication Engineering, Dalian University of Technology, Dalian, China, 116024. (e-mails: xfgong@dlut.edu.cn, qhlin@dlut.edu.cn).

Feng-Yu Cong is with the Department of Biomedical Engineering, Dalian University of Technology, Dalian, China, 116024 (e-mail: cong@dlut.edu.cn).

Lieven De Lathauwer is with the STADIUS Center for Dynamical Systems, Signal Processing and Data Analytics; Department of Electrical Engineering (ESAT), KU Leuven, BE-3001 Leuven, Belgium; and Group Science, Engineering and Technology, KU Leuven, Kulak 8500 Kortrijk, Belgium (e-mail: Lieven.DeLathauwer@kuleuven.be).

C-CPD approaches in [47]–[49] are limited to the case of two coupled tensor datasets. The approaches in [50]–[58] based on either hard or soft coupling structure are mainly used to fuse multiple datasets which have tensor form, while the general multi-set data formulation in J-BSS is in matrix form. The T-PICA variants [42]–[44] assume that either the mixing matrices or the sources for different datasets are the same, and thus may not work well in applications where this crucial assumption is not satisfied. In fact, as will be shown later, by using second-order statistics of the multi-set signal, we can obtain a set of tensors admitting a specific double coupled CPD (DC-CPD) with coupling in two modes, possibly with a conjugation structure for complex-valued problems (an illustration is given in Fig.1). The coupling factor matrices of this specific DC-CPD may not have full column rank. We note that the GJD formulation in [36]–[39] is similar to DC-CPD, but is limited to the case where the coupling factor matrices have full column rank. When revising this paper, we noticed that an optimization based DC-CPD algorithm has recently been proposed for real-valued underdetermined J-BSS [60]. However, the algebraic aspects of DC-CPD were not studied in [60]. As such, although some relevant works have recently appeared in the literature, the specific but important DC-CPD problem, including uniqueness conditions and computation by linear algebraic means, has not yet been thoroughly studied to the best of our knowledge. It will be addressed in this paper, both from a theoretical and a practical perspective.

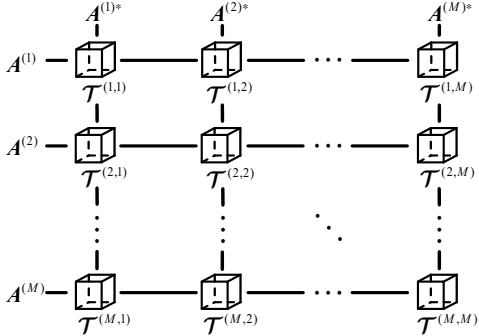

Fig.1. Illustration of DC-CPD. The tensors are placed at different nodes of a grid according to their indices. The tensor at node $(m, n)$ admits $\mathcal{T}^{(m,n)} = [\![A^{(m)}, A^{(n)*}, C^{(m,n)}]\!]_R$, $m, n = 1,\ldots,M$. Tensors in the $m$th row of the grid are coupled in the first mode by $A^{(m)}$. Tensors in the $m$th column of the grid are coupled in the second mode by $A^{(m)*}$.

In Section II, we give the definition of DC-CPD and explain how second-order statistics can be used to obtain a DC-CPD from the matrix data in J-BSS. In Section III, we present an algebraic algorithm based on a new coupled rank-1 detection mapping. We will show that it is possible to compute an exact DC-CPD by means of conventional linear algebra, essentially by solving overdetermined sets of linear equations, low-rank matrix approximation and generalized eigenvalue decomposition (GEVD), even in underdetermined cases where the number of sources exceeds the number of observation channels. In addition, we obtain deterministic uniqueness conditions for DC-CPD. The main result is *Theorem 2*. In Section IV, we introduce optimization based DC-CPD methods, including alternating least squares (ALS) and structured data fusion (SDF) [64] based methods. In Section V, we discuss several theoretical and practical aspects of DC-CPD, including generic uniqueness conditions, to provide further insight. In Section VI, we provide experimental results to demonstrate the performance of DC-CPD, in comparison with CPD and other existing J-BSS algorithms. Section VII concludes this paper.

*Notation:* Vectors, matrices and tensors are denoted by lowercase boldface, uppercase boldface and uppercase calligraphic letters, respectively. The $r$th column vector and the $(i, j)$th entry of $A$ are denoted by $a_r$ and $a_{i,j}$, respectively. The identity matrix and all-zero vectors are denoted by $I_M \in \mathbb{C}^{M \times M}$ and $0_M$ respectively. The subscripts are omitted if there is no ambiguity. The null space of a matrix $M$ is denoted as $\ker(M)$. The dimensionality of a vector space $\mathfrak{I}$ is denoted as $\dim(\mathfrak{I})$. Transpose, conjugate, conjugated transpose, Moore-Penrose pseudo-inverse, Frobenius norm and matrix determinant are denoted as $(\cdot)^T$, $(\cdot)^*$, $(\cdot)^H$, $(\cdot)^\dagger$, $\|\cdot\|_F$, and $|\cdot|$, respectively.

The symbols '$\otimes$', '$\odot$' and '$\otimes$' denote Kronecker product, Khatri-Rao product, and tensor outer product, respectively, defined as:

$$A \otimes B \triangleq \begin{bmatrix} a_{11}B & a_{12}B & \cdots \\ a_{21}B & a_{22}B & \cdots \\ \vdots & \vdots & \ddots \end{bmatrix}, \quad \begin{array}{l} A \odot B \triangleq [a_1 \otimes b_1, a_2 \otimes b_2, \cdots] \\ (a \otimes b \otimes c)_{i,j,k} \triangleq a_i b_j c_k \end{array}.$$

Mathematical expectation is denoted as $\mathrm{E}\{\cdot\}$. MATLAB notation will be used to denote submatrices of a tensor. For instance, we use $(\mathcal{T})_{(:,:,k)}$ or $T_{(:,:,k)}$ to denote the frontal slice of tensor $\mathcal{T}$ obtained by fixing the third index to $k$. A polyadic decomposition (PD) of $\mathcal{T}$ expresses $\mathcal{T}$ as the sum of rank-1 terms:

$$\mathcal{T} = [\![A, B, C]\!]_R \triangleq \sum_{r=1}^R a_r \otimes b_r \otimes c_r \in \mathbb{C}^{I \times J \times K}, \quad (1)$$

where $A \triangleq [a_1,\ldots,a_R] \in \mathbb{C}^{I \times R}$, $B \triangleq [b_1,\ldots,b_R] \in \mathbb{C}^{J \times R}$, and $C \triangleq [c_1,\ldots,c_R] \in \mathbb{C}^{K \times R}$. We call (1) a canonical PD (CPD) if $R$ is minimal.

For an $N$th-order tensor $\mathcal{T} \in \mathbb{C}^{I_1 \times \cdots \times I_N}$, $\mathrm{vec}(\mathcal{T}) \in \mathbb{C}^{I_1 \cdots I_N}$ denotes the vector representation of $\mathcal{T}$: $[\mathrm{vec}(\mathcal{T})]_{\tilde{i}} \triangleq t_{i_1,\ldots,i_N}$, with $\tilde{i} = \sum_{n=1}^N (i_n - 1) \prod_{m=1}^{N-n} I_m + 1$, while $\mathrm{unvec}(\cdot)$ performs the inverse. The mode-$i$ matrix representation of a third-order tensor $\mathcal{T} \in \mathbb{C}^{I \times J \times K}$ is denoted as $T_i$, $i = 1, 2, 3$, and defined by:

$$(T_1)_{(j-1)K+k,i} = (T_2)_{(i-1)K+k,j} = (T_3)_{(i-1)J+j,i} = t_{i,j,k}.$$

We define $\mathrm{Ten}(T, [I, J, K]) = \mathcal{T}$ as the operation to reshape an $IJ \times K$ matrix $T$ into a third-order tensor $\mathcal{T}$ of size $I \times J \times K$, such that $t_{i,j,k} = T((i-1)J + j, k)$.

## II. PROBLEM FORMULATION

### A. DC-CPD

We say that a set of tensors with varying indices $m$ and $n$, $\{\mathcal{T}^{(m,n)} \in \mathbb{C}^{N_m \times N_n \times T}, m, n = 1,\ldots,M\}$, admits an $R$-term double coupled PD (DC-PD) if each tensor $\mathcal{T}^{(m,n)}$ can be written as:

$$\mathcal{T}^{(m,n)} = \sum_{r=1}^R a_r^{(m)} \otimes a_r^{(n)*} \otimes c_r^{(m,n)} = [\![A^{(m)}, A^{(n)*}, C^{(m,n)}]\!]_R, \quad (2)$$

with factor matrices $A^{(m)} \triangleq [a_1^{(m)},\ldots,a_R^{(m)}] \in \mathbb{C}^{N_m \times R}$ and $C^{(m,n)} \triangleq [c_1^{(m,n)},\ldots,c_R^{(m,n)}] \in \mathbb{C}^{T \times R}$. The rank-1 terms with fixed $r$ and varying $m$ and $n$, $\{a_r^{(m)} \otimes a_r^{(n)*} \otimes c_r^{(m,n)}, m,n = 1,\ldots,M\}$, are together denoted as a double coupled rank-1 term. If the number of double coupled rank-1 terms in (2), $R$, is minimal, then the decomposition (2) is denoted as the DC-CPD of $\{\mathcal{T}^{(m,n)}\}$, and $R$ is denoted as the double coupled rank of $\{\mathcal{T}^{(m,n)}\}$.

In practice, due to noise and model errors, equation (2) is usually only approximate. In addition to the exact DC-CPD (2), we will consider the approximate DC-CPD, where the problem is to find factor matrices $A^{(m)} \in \mathbb{C}^{N_m \times R}, C^{(m,n)} \in \mathbb{C}^{T \times R}, m, n = 1,...,M$, such that the overall approximation error $\eta$ is minimized in the least squares (LS) sense:

$$\{\tilde{A}^{(m)}, \tilde{C}^{(m,n)}, m, n \in [1, M]\} = \underset{\{A^{(m)}, C^{(m,n)}\}}{\arg\min} \eta, \quad (3)$$

where

$$\eta = \sum_{m,n} \left\| \mathcal{T}^{(m,n)} - [\![A^{(m)}, A^{(n)*}, C^{(m,n)}]\!]_R \right\|_F^2. \quad (4)$$

The term "double coupled" is used to indicate that each PD shares factor matrices with other PDs in the first two modes. That is to say, if we place the tensors $\mathcal{T}^{(m,n)}$ at different nodes of a grid according to their indices (see Fig.1), the tensors in the same "row" (e.g. the $m$th row) are coupled in the first mode (by $A^{(m)}$), and all the tensors in the same "column" (e.g. the $n$th column) are coupled in the second mode (by $A^{(n)*}$).

The double coupled rank-1 terms $\{a_r^{(m)} \otimes a_r^{(n)*} \otimes c_r^{(m,n)}, m, n = 1,...,M\}$ in (2) can be arbitrarily permuted and the vectors $a_r^{(m)}, a_r^{(n)*}$ and $c_r^{(m,n)}$ with fixed $r, m, n$ can be arbitrarily scaled provided the overall double coupled rank-1 term remains the same. We say that the DC-CPD is unique when it is only subject to these trivial indeterminacies.

### B. Formulation of J-BSS into a DC-CPD

J-BSS usually assumes the following multi-set data model:

$$x^{(m)}(t) = A^{(m)} s^{(m)}(t), \quad m = 1,...,M, \quad (5)$$

where $x^{(m)}(t) \in \mathbb{C}^{N_m}$, $s^{(m)}(t) \in \mathbb{C}^R$ denote the mixture and source at time instant $t$, respectively, and $A^{(m)} \in \mathbb{C}^{N_m \times R}$ is the mixing matrix for the $m$th dataset. In practice, there is also a noise term. Here we omit it for convenience. $N_m$, $R$ and $M$ are the number of observation channels in the $m$th dataset, the number of sources and datasets, respectively. J-BSS aims at estimating the mixing matrices $A^{(m)}$ and/or sources $s^{(m)}(t)$ for $m = 1,...,M$. We call a J-BSS underdetermined if one or more of the mixing matrices $A^{(1)},...,A^{(M)}$ do not have full column rank, which is necessarily the case if they have more columns than rows. Otherwise, we call it overdetermined.

With the multi-set data model (5), we additionally assume that $s_r^{(m)}(t)$ and $s_u^{(n)}(t)$ are independent for $1 \le r \ne u \le R$ (intra-set independence) and dependent for $1 \le r = u \le R$ (inter-set dependence), regardless of the value of $m$ and $n$. The idea is now to obtain a set of tensors by stacking second-order cross-covariance matrices. We assume that the sources are temporally nonstationary with zero mean and unit variance. Then we obtain a set of cross-covariance tensors as:

$$\begin{aligned}\left(\mathcal{T}^{(m,n)}\right)_{(:,:,t)} &= \mathrm{E}\{x^{(m)}(t)[x^{(n)}(t)]^H\} \\ &= A^{(m)} \mathrm{E}\{s^{(m)}(t)[s^{(n)}(t)]^H\} A^{(n)H},\end{aligned} \quad (6)$$

where $\mathcal{T}^{(m,n)} \in \mathbb{C}^{N_m \times N_n \times T}$, $1 \le m, n \le M$. We note that the frontal slice $(\mathcal{T}^{(m,n)})_{(:,:,t)}$ is the cross-covariance matrix between the $m$th and $n$th datasets at time instant $t$, and that $T$ denotes the number of time frames for which such a cross-covariance is computed. Noting that $\mathrm{E}\{s^{(m)}(t) \cdot [s^{(n)}(t)]^H\}$ is diagonal under the assumption of intra-set independence and inter-set dependence, we can rewrite (6) as follows:

$$\mathcal{T}^{(m,n)} = \sum_{r=1}^R a_r^{(m)} \otimes a_r^{(n)*} \otimes c_r^{(m,n)} = [\![A^{(m)}, A^{(n)*}, C^{(m,n)}]\!]_R, \quad (7)$$

where $c_r^{(m,n)}(t) = \mathrm{E}\{s_r^{(m)}(t)[s_r^{(n)}(t)]^H\}$, and $C^{(m,n)} \triangleq [c_1^{(m,n)},..., c_R^{(m,n)}]$. Note that the set of tensors $\{\mathcal{T}^{(m,n)}, m, n = 1,...,M\}$, admits a DC-CPD (2), and our goal is to decompose the tensors $\{\mathcal{T}^{(m,n)}\}$ into a minimal number of double coupled rank-1 terms, to recover the mixing matrices $A^{(m)}$.

So far, we have shown how, via the computation of second-order statistics, the stochastic multi-set signals in (5) are transformed into a set of tensors, the joint decomposition of which takes the form of a DC-CPD (2). This method fits in the tensorization framework proposed in [61]. In line with the discussion in [61], the inherent intra-set independence and inter-set dependence of sources are key in the conversion, and have in fact already been used in existing practical multi-set signal processing applications, not necessarily in the J-BSS framework sensu stricto. For example, for linear convolutive mixtures that are transformed to the frequency domain, the cross-correlations of components at neighboring frequency bins are intensively exploited to solve the permutation misalignment problem [30], [39], [47], [63]. The above assumptions on multiset statistics are also used in group analysis of biomedical data [29], [31]–[39] where the multiple datasets may refer to data of multiple modalities (e.g. fMRI, MEG, EEG) collected from a single subject under equal conditions, or data collected from multiple subjects, possibly using a single modality but acquired under the same conditions.

In the rest of the paper, we assume for convenience that $N_1 = N_2 = \cdots = N_M = N$, so that all the tensors $\mathcal{T}^{(m,n)}$ have identical size $N \times N \times T$. The derivation and results remain similar in the general case where $N_1,...,N_M$ are not the same.

In addition, we assume that the DC-CPD has a conjugated symmetric structure: $\mathcal{T}^{(m,n)} = \mathrm{perm}_{(2,1,3)}(\mathcal{T}^{(n,m)*})$, where $\mathrm{perm}_{(2,1,3)}(\cdot)$ permutes the first and second indices of its entry. We note that this symmetry is present for the DC-CPD derived via (6). In cases where this symmetry is not readily present, we are able to create it by tensor concatenation. The details are given in *Section I* of the *Supplementary materials*.

Because of the conjugated symmetry in DC-CPD, it suffices to consider the following set $\Upsilon^{(m)}$ of tensors to take into account all occurrences of $A^{(m)}$, for fixed $m$:

$$\Upsilon^{(m)} \triangleq \left\{ \mathcal{T}^{(m,n)} = [\![A^{(m)}, A^{(n)*}, C^{(m,n)}]\!]_R, n = 1,...,M \right\}. \quad (8)$$

Note that in (8), seen as C-CPD, occurrences of $A^{(m)*}$ in the second mode are taken care of automatically because of the conjugated symmetry.

The DC-CPD (2) differs from existing C-CPD works in the following aspects: (i) DC-CPD has a double coupling structure in the first two modes while C-CPD in [45]–[58] consider coupling in a single mode; (ii) the common factor matrix in DC-CPD corresponds to the mixing matrix in J-BSS (5), and does not have full column rank if the J-BSS problem is underdetermined. The existing C-CPD works all assume that the common factor matrix has full column rank. In [59] it is assumed that at least one of the common factor matrices in a DC-CPD has full column rank. In what follows, we call a DC-CPD overdetermined if all the factor matrices $A^{(1)},...,A^{(M)}$ have full column rank. The DC-CPD is called underdetermined if it is not the case. Underdetermined DC-CPD may result from underdetermined J-BSS problems, in which one or more of the mixing matrices $A^{(1)},...,A^{(M)}$ have more columns than rows.

## III. ALGEBRAIC ALGORITHM

In this section, we introduce an algebraic algorithm for DC-CPD. We formally call an algorithm algebraic if it relies only on arithmetic operations, overdetermined sets of linear equations, matrix singular value decomposition (SVD) and generalized eigenvalue decomposition (GEVD). In particular, an algebraic computation does not involve a numerical optimization in which the cost function may have multiple local optima, or of which the convergence is not guaranteed.

If the DC-CPD (2) is *overdetermined*, we immediately have an algebraic solution. More precisely, any of the CPDs in (2) can be computed by GEVD [62]. This yields us the factor matrices $A^{(m)}$ and $A^{(n)}$ up to trivial scaling and permutation indeterminacies, for any choice of $m$ and $n$. Assume for instance that $A^{(1)}$ has been found, then we can find all the remaining $A^{(n)}$, $n = 2,...,M$, from the CPD of $\mathcal{T}^{(1,n)}$ for varying $n$. In fact, since in these CPDs the factor matrix $A^{(1)}$ is already known, the other factor matrices are obtained by mere rank-1 approximation of the matricized columns of [22]:

$$T_1^{(1,n)}(A^{(1)T})^\dagger = A^{(n)*} \odot C^{(1,n)}. \quad (9)$$

Indeed, each column of $A^{(n)*} \odot C^{(1,n)}$ is a vectorized rank-1 matrix:

$$\text{unvec}(q_{1,r}^{(1,n)}) = a_r^{(n)*} \cdot c_r^{(1,n)T}, \quad r=1,...,R, \quad (10)$$

where vector $q_{1,r}^{(1,n)}$ is the $r$th column of $T_1^{(1,n)}(A^{(1)T})^\dagger$. Continuing this way, all the factor matrices $A^{(m)}$ can be determined, $m = 1,...,M$. The overdetermined DC-CPD can also be computed by GJD algorithms proposed in [37]–[39], noting that GJD are essentially equivalent to DC-CPD in the overdetermined case.

If the DC-CPD (2) is *underdetermined*, it is still possible to derive an algebraic algorithm, which is the main concern of our work.

### A. Basic assumptions

We assume both $A^{(m)} \odot A^{(n)*}$ and $C^{(m,n)}$ have full column rank, for all the indices $m, n$. These assumptions together imply that the rank of tensor $\mathcal{T}^{(m,n)}$ is equal to the rank of the mode-3 matrix representation $T_3^{(m,n)}$. This is of practical use in the sense that the number of sources can be determined by checking the number of significant singular values of $T_3^{(m,n)}$. We note that DC-CPD (7) is not unique if $A^{(m)} \odot A^{(n)*}$ does not have full column rank. In fact, in this case a decomposition in a smaller number of terms is possible. For instance, if $a_R^{(m)} \otimes a_R^{(n)*} = \sum_{r=1}^{R-1} \alpha_r^{(m,n)} a_r^{(m)} \otimes a_r^{(n)*}$, then $\mathcal{T}^{(m,n)} = \sum_{r=1}^{R-1} a_r^{(m)} \otimes a_r^{(n)*} \otimes (c_r^{(m,n)} + \alpha_r^{(m,n)} c_R^{(m,n)})$. On the other hand, if $C^{(m,n)}$ does not have full column rank, decomposition (2) may still be unique (for instance, algebraic algorithms for CPD have been derived for cases where none of the factor matrices has full column rank [12], [13]), while the rank of tensor $\mathcal{T}^{(m,n)}$ is not equal to the rank of the mode-3 matrix representation $T_3^{(m,n)}$. Here we do not consider this more complicated case. In addition, we make the notational assumption that the factor matrices $C^{(m,n)}$ in the third mode have size $R \times R$. In practice, this can always be achieved by a classical dimensionality reduction step: if the columns of a matrix $U^{(m,n)}$ form an orthonormal basis of the row space of $T_3^{(m,n)}$, then $T_3^{(m,n)}U^{(m,n)}$ has $R$ columns, the tensor $\text{Ten}(T_3^{(m,n)}U^{(m,n)}, I, I, R)$ has reduced dimensionality $R$ in the third mode, and its factor matrices are $A^{(m)}$, $A^{(n)*}$, and $U^{(m,n)T}C^{(m,n)} \in \mathbb{C}^{R \times R}$.

### B. Review of algebraic CPD algorithm

The algebraic DC-CPD algorithm fits in the framework of the algebraic algorithms in [10], [12], [13], [19], [20], [51], [52], which, by the use of a rank-1 or rank-$R$ detection mapping, convert a possibly underdetermined (coupled) CPD into an overdetermined CPD. To make the derivation of our algebraic DC-CPD algorithm more accessible, we first give a high-level summary of the algebraic algorithm for underdetermined CPD [10]. For more details, the reader is referred to [10].

First, the rank-1 detection mapping $\Phi^{(R1)}: (X,Y) \in \mathbb{C}^{I \times J} \times \mathbb{C}^{I \times J} \to \Phi^{(R1)}(X,Y) \in \mathbb{C}^{I \times I \times J \times J}$ is defined as:

$$\left[\Phi^{(R1)}(X,Y)\right]_{i,j,g,h} \triangleq x_{i,g}y_{j,h} + y_{i,g}x_{j,h} - x_{i,h}y_{j,g} - y_{i,h}x_{j,g}. \quad (11)$$

Three main properties of this rank-1 detection mapping are: (1) it is bilinear in its arguments: $\Phi^{(R1)}(\sum_p \alpha_p X_p, \sum_q \beta_q Y_q) = \sum_p \sum_q \alpha_p \beta_q \Phi^{(R1)}(X_p, Y_q)$; (2) for a non-zero matrix $X$, $\Phi^{(R1)}(X,X)$ is a zero tensor if and only if $X$ is a rank-1 matrix; and (3) the rank-1 detection mapping is symmetric in its arguments $\Phi^{(R1)}(X,Y) = \Phi^{(R1)}(Y,X)$.

To compute the CPD of a tensor $\mathcal{T} = [\![A, A', C]\!]_R$, where $A \in \mathbb{C}^{I \times R}, A' \in \mathbb{C}^{J \times R}$, and $C \in \mathbb{C}^{R \times R}$, the algebraic CPD algorithm in [10] consists of the following main steps (note that only $C$ is required to have full column rank, $A$ and/or $A'$ can be rank-deficient, and we can allow $I < R$ and/or $J < R$):

(*i*) Apply the rank-1 detection mapping to the $s$th and $u$th frontal slices of $\mathcal{T}$ to construct a tensor $\Phi^{(R1)}(T_{(:,:,s)}, T_{(:,:,u)})$, and reshape it into a vector of length $I^2 J^2$. We have $R(R-1)/2$ such vectors for varying $s$ and $u$, $1 \leq s < u \leq R$. We stack these vectors into the columns of an $I^2 J^2 \times R(R-1)/2$ matrix $\Gamma$.

(*ii*) We make the assumption that tensors $\Phi^{(R1)}(a_t a_t'^T, a_r a_r'^T)$, are linearly independent for $1 \leq t < r \leq R$. This assumption is very mild. It can be shown that generically it is satisfied as long as $R(R-1) \leq IJ(I-1)(J-1)/2$. Under this assumption we have: $\dim(\ker(\Gamma)) = R$, and the $R$ basis vectors $w_1,...,w_R$ in $\ker(\Gamma)$ can be reshaped into a tensor that admits an overdetermined CPD: $\mathcal{W} = [\![B, B, F]\!]_R$, where factor matrix $B = \tilde{C}^{-T}$, with $\tilde{C}$ equivalent to $C$ up to scaling and permutation ambiguities.

(*iii*) As the CPD $\mathcal{W} = [\![B, B, F]\!]_R$ is overdetermined, it can be computed algebraically via GEVD, which makes use of two frontal slices of $\mathcal{W}$. We can also compute it via matrix simultaneous diagonalization (SD) that makes use of all the frontal slices. Optimization based CPD algorithms such as alternating least squares (ALS), or nonlinear least squares (NLS) can be used to maximize the overall fit. Once $B$ has been computed, we immediately obtain $\tilde{C} = B^{-T}$, and the remaining factor matrices $A$ and $A'$ can be recovered using the CPD structure of the data tensor.

Next, in *Subsection III.C*, we will first propose a variant of the rank-1 detection mapping (11), namely the coupled rank-1 detection mapping (12). Then, in *Subsection III.D*, we will derive the algebraic DC-CPD algorithm using the coupled rank-1 detection mapping. The derivation follows a similar line of thought as for its CPD counterpart, as will be shown later.

### C. Coupled rank-1 detection mapping

We give the following two definitions and a theorem.

*Definition 1:* Two matrices $X^{(1)} \in \mathbb{C}^{N \times P}, X^{(2)} \in \mathbb{C}^{N \times Q}$ are said to be coupled rank-1 matrices if they are both rank-1 and have the same column space.

*Definition 2:* The coupled rank-1 detection mapping $\Phi$: $(X^{(1)}, X^{(2)}) \in \mathbb{C}^{N \times P} \times \mathbb{C}^{N \times Q} \to \Phi(X^{(1)}, X^{(2)}) \in \mathbb{C}^{N \times N \times P \times Q}$ is defined as:

$$\left[\Phi(X^{(1)}, X^{(2)})\right]_{i,j,p,q} \triangleq \begin{vmatrix} x_{i,p}^{(1)} & x_{i,q}^{(2)} \\ x_{j,p}^{(1)} & x_{j,q}^{(2)} \end{vmatrix} = x_{i,p}^{(1)} x_{j,q}^{(2)} - x_{j,p}^{(1)} x_{i,q}^{(2)}. \quad (12)$$

We define $\psi(X^{(1)}, X^{(2)})$ of length $N^2 PQ$ as the vector representation of $\Phi(X^{(1)}, X^{(2)})$:

$$\psi(X^{(1)}, X^{(2)}) \triangleq \text{vec}\left(\Phi(X^{(1)}, X^{(2)})\right). \quad (13)$$

*Theorem 1:* For two nonzero matrices $X^{(1)} \in \mathbb{C}^{N \times P}, X^{(2)} \in \mathbb{C}^{N \times Q}$, the vector $\psi(X^{(1)}, X^{(2)})$ is a zero vector if and only if its arguments $X^{(1)}, X^{(2)}$ are two coupled rank-1 matrices.

We can easily prove *Theorem 1*, according to (12), (13), and the definition of coupled rank-1 matrices. We note that the above definitions and theorem generalize the rank-1 detection mapping and relevant results [10] to the coupled case.

Next we explain how to compute an underdetermined DC-CPD via the coupled rank-1 detection mapping.

### D. DC-CPD via coupled rank-1 detection mapping

We first formulate a theorem that provides deterministic conditions for which DC-CPD is unique and for which DC-CPD can be calculated algebraically.

*Theorem 2:* Let $\mathcal{T}^{(m,l)} = [\![ A^{(m)}, A^{(l)*}, C^{(m,l)} ]\!]_R$, where $l \in \{g, h\}$. Assume that $C^{(m,g)}, C^{(m,h)}$ have full column rank for all $1 \le g < h \le M, m = 1, ..., M$, and that $\psi(a_t^{(m)} a_t^{(g)H}, a_r^{(m)} a_r^{(h)H}) = (a_t^{(m)} \otimes a_r^{(m)} - a_r^{(m)} \otimes a_t^{(m)}) \otimes a_t^{(g)*} \otimes a_r^{(h)*}$ are linearly independent for $1 \le t \ne r \le R$. Then we have:

- Tensors $\{\mathcal{T}^{(m,n)}, m, n = 1, ..., M\}$ admit a DC-CPD, i.e. they consist of the sum of $R$ double coupled rank-1 terms, and the number of terms cannot be reduced.
- The DC-CPD is unique.
- The DC-CPD can be calculated algebraically.

Next we derive the algebraic DC-CPD algorithm. Note that the derivation provides a constructive proof of *Theorem 2*. The main idea is to transform the original underdetermined DC-CPD into an overdetermined DC-CPD through the use of the coupled rank-1 detection mapping (12). All the factor matrices involved in this overdetermined DC-CPD have full column rank, and thus can be calculated algebraically via GEVD. If the original DC-CPD is already overdetermined, then the rank-1 detection mapping need not be used, and the factor matrices can be directly computed via GEVD or SD.

(*i*) *Construct matrix $\Gamma^{(m,g,h)}$ via coupled rank-1 detection.*

For each pair of tensors $\mathcal{T}^{(m,g)}, \mathcal{T}^{(m,h)} \in \Upsilon^{(m)}$, we use the operation $\Phi(\cdot)$ on the $s$th frontal slice of the former, and the $u$th frontal slice of the latter, $1 \le s, u \le R$:

$$\left[\Phi(T_{(:,:,s)}^{(m,g)}, T_{(:,:,u)}^{(m,h)})\right]_{i,j,p,q} = t_{i,p,s}^{(m,g)} t_{j,q,u}^{(m,h)} - t_{j,p,s}^{(m,g)} t_{i,q,u}^{(m,h)}. \quad (14)$$

We vectorize $\Phi(T_{(:,:,s)}^{(m,g)}, T_{(:,:,u)}^{(m,h)})$ into vectors $\psi(T_{(:,:,s)}^{(m,g)}, T_{(:,:,u)}^{(m,h)})$, and stack these vectors into the columns of an $N^4 \times R^2$ matrix $\Gamma^{(m,g,h)}$. Due to the bilinearity of $\psi$, and by making use of the property that $\psi(a_t^{(m)} a_t^{(g)H}, a_r^{(m)} a_r^{(h)H}) = 0$ when $t = r$ (*Theorem 1*), we have the following result after a short derivation given in *Section II* of the *Supplementary materials*:

$$\Gamma^{(m,g,h)} = \Phi_{g,h}^{(m)} \cdot P_{g,h}^{(m)T}, \quad (15)$$

where the $N^4 \times (R^2 - R)$ matrix $\Phi_{g,h}^{(m)}$ holds the vectors $\psi(a_t^{(m)} a_t^{(g)H}, a_r^{(m)} a_r^{(h)H})$ as its columns, and the $R^2 \times (R^2 - R)$ matrix $P_{g,h}^{(m)}$ holds $c_t^{(m,g)} \otimes c_r^{(m,h)}$ as its columns. The columns are indexed by $(t, r)$, where $1 \le t \ne r \le R$.

Note that $\Gamma^{(m,g,h)}$ is a DC-CPD variant of the matrix $\Gamma$ for CPD in step (*i*) of *Subsection III.B*.

(*ii*) *Construct tensor $\mathcal{W}^{(m,g,h)}$ in $\ker(\Gamma^{(m,g,h)})$. For varying m, g, h, these tensors admit an overdetermined DC-CPD.*

We give a theorem implying that the basis vectors in $\ker(\Gamma^{(m,g,h)})$ can be reshaped into a tensor that admits an overdetermined third-order CPD, from which the factor matrices $C^{(m,n)}$ can be obtained.

*Theorem 3:* Let $\mathcal{T}^{(m,l)} = [\![ A^{(m)}, A^{(l)*}, C^{(m,l)} ]\!]_R \in \Upsilon^{(m)}$, where $l \in \{g, h\}$, $1 \le g < h \le M$, and assume that $\Phi_{g,h}^{(m)}$ in (15) and $C^{(m,g)}, C^{(m,h)}$ have full column rank for fixed $(m, g, h)$, then we have:

(i) $\ker(\Gamma^{(m,g,h)}) = \ker(P_{g,h}^{(m)T})$.
(ii) $\dim(\ker(\Gamma^{(m,g,h)})) = R$.
(iii) The basis vectors $w_r^{(m,g,h)}$ in $\ker(\Gamma^{(m,g,h)})$, $r = 1, ..., R$, can be written as linear combinations of the vectors $b_u^{(m,g)} \otimes b_u^{(m,h)}$, $u = 1, ..., R$ :

$$w_r^{(m,g,h)} = \sum_{u=1}^{R} f_{r,u}^{(m,g,h)} \cdot b_u^{(m,g)} \otimes b_u^{(m,h)} \in \mathbb{C}^{R^2}, \quad (16)$$

where $[b_1^{(m,l)}, ..., b_R^{(m,l)}] = B^{(m,l)} = (\tilde{C}^{(m,l)})^{-T}$, $\tilde{C}^{(m,l)} \triangleq C^{(m,l)} D_C^{(m,l)} \Pi$, $D_C^{(m,l)}$ is a diagonal matrix, $l \in \{g, h\}$, and $\Pi$ is a permutation matrix that is common for $C^{(m,g)}$ and $C^{(m,h)}$. That is to say, the permutation between the columns of $\tilde{C}^{(m,g)}$ and $C^{(m,g)}$ is necessarily the same as that between the columns of $\tilde{C}^{(m,h)}$ and $C^{(m,h)}$, due to the fact that in the decomposition of $\mathcal{T}^{(m,g)}$ and $\mathcal{T}^{(m,h)}$ the factor matrix $A^{(m)}$ is shared.

We give the proof of this theorem in the *Appendix*. The theorem provides the following key results for the algebraic algorithm: (a) the dimension of the null space of $\Gamma^{(m,g,h)}$ reveals the number of sources; (b) via (16), the basis vectors in the null space of $\Gamma^{(m,g,h)}$ are explicitly linked to the inverse of $C^{(m,l)T}$ up to trivial indeterminacies.

Equation (16) shows that we can reshape the $R^2 \times R$ matrix $W^{(m,g,h)} \triangleq [w_1^{(m,g,h)}, ..., w_R^{(m,g,h)}]$ into a tensor: $\mathcal{W}^{(m,g,h)} \triangleq \text{Ten}(W^{(m,g,h)}, [R, R, R])$, which admits a third-order CPD, for

fixed $(m, g, h)$:

$$\mathcal{W}^{(m,g,h)} = [\![\boldsymbol{B}^{(m,g)}, \boldsymbol{B}^{(m,h)}, \boldsymbol{F}^{(m,g,h)}]\!]_R, \quad (17)$$

where $\boldsymbol{F}^{(m,g,h)} \in \mathbb{C}^{R \times R}$ holds $f_{r,u}^{(m,g,h)}$ as its $(r, u)$th entry. Recall that $\boldsymbol{C}^{(m,g)}, \boldsymbol{C}^{(m,h)}$ have full column rank by assumption, and thus $\boldsymbol{B}^{(m,g)}, \boldsymbol{B}^{(m,h)}$ also have full column rank. In addition, the matrix $\boldsymbol{W}^{(m,g,h)}$ has full column rank, as its columns are basis vectors of $\ker(\boldsymbol{\Gamma}^{(m,p,q)})$. According to (17) we have $\boldsymbol{F}^{(m,g,h)T} = (\boldsymbol{B}^{(m,g)} \odot \boldsymbol{B}^{(m,h)})^\dagger \boldsymbol{W}^{(m,g,h)}$, and thus $\boldsymbol{F}^{(m,p,q)}$ also has full rank. As such, all the factor matrices of $\mathcal{W}^{(m,p,q)}$ have full column rank, and $\mathcal{W}^{(m,p,q)}$ admits a CPD that is overdetermined.

In (14)–(17), we have used the coupled rank-1 detection mapping (12) to convert the decomposition of a pair of coupled tensors into a non-symmetric overdetermined third-order CPD. For the overall DC-CPD (8), we perform the same procedure for all pairs of coupled tensors and obtain a set of overdetermined CPDs. With varying indices $(m, g, h)$, $m = 1,\ldots,M$, $1 \le g < h \le M$, the CPDs (17) together form an overdetermined third-order DC-CPD with coupling in the first two modes. Recall that the factor matrices of the overdetermined DC-CPD, $\boldsymbol{B}^{(m,g)}$ and $\boldsymbol{B}^{(m,h)}$, are up to trivial indeterminacies equivalent to the inverse of $\boldsymbol{C}^{(m,g)T}$ and $\boldsymbol{C}^{(m,h)T}$, respectively, and thus computing the overdetermined DC-CPD of $\mathcal{W}^{(m,p,q)}$ yields estimates of the factor matrices $\boldsymbol{C}^{(m,l)}$. Subsequently, the estimates of the factor matrices $\boldsymbol{A}^{(m)}$ will be obtained.

Note that the derivation in this subsection is in analogy with that in CPD step (*ii*) in *Subsection III.B*.

(*iii*) Solve the overdetermined DC-CPD (17), and calculate factor matrices $\boldsymbol{A}^{(1)},\ldots,\boldsymbol{A}^{(M)}$.

As the new DC-CPD (17) is overdetermined, it admits an algebraic solution. There are different ways to compute this solution. First, as all the factor matrices have full column rank, any of the CPDs (17) can be computed by GEVD [62]. This yields us the factor matrices $\boldsymbol{B}^{(m,g)}$ and $\boldsymbol{B}^{(m,h)}$ up to trivial scaling and permutation indeterminacies, for any choice of $m$, $g$ and $h$. Assume for instance that $\boldsymbol{B}^{(m,1)}$ has been found, then we can find all the remaining $\boldsymbol{B}^{(m,h)}$ and $\boldsymbol{F}^{(m,1,h)}$ from the CPD of $\mathcal{W}^{(m,1,h)}$ for fixed $m$ and varying $h$. In fact, since in these CPDs the factor matrix $\boldsymbol{B}^{(m,1)}$ is already known, the other factor matrices follow from rank-1 approximation of the matricized columns of [22]:

$$\boldsymbol{W}_1^{(m,1,h)} \boldsymbol{B}^{(m,1)-T} = \boldsymbol{B}^{(m,h)} \odot \boldsymbol{F}^{(m,1,h)}. \quad (18)$$

Indeed, each column of $\boldsymbol{B}^{(m,h)} \odot \boldsymbol{F}^{(m,1,h)}$ is a vectorized rank-1 matrix:

$$\text{unvec}(\boldsymbol{q}_{1,r}^{(m,1,h)}) = \boldsymbol{b}_r^{(m,h)} \cdot \boldsymbol{f}_r^{(m,1,h)T}, \quad r = 1,\ldots,R, \quad (19)$$

where vector $\boldsymbol{q}_{1,r}^{(m,1,h)}$ of length $R^2$ is the $r$th column of $\boldsymbol{W}_1^{(m,1,h)} \boldsymbol{B}^{(m,1)-T}$. Continuing this way, all the factor matrices $\boldsymbol{B}^{(m,n)}$ can be determined, $m, n = 1,\ldots,M$.

Besides the above-mentioned GEVD based scheme, which makes use of two frontal slices of the data tensor, we may consider computing any of the CPDs (17) via SD, and then follow a similar strategy as the one above to solve the DC-CPD. We may also consider computing the DC-CPD of all the tensors $\mathcal{W}^{(m,g,h)}$ in (17) simultaneously via an optimization based algorithm, while taking the coupling structure into account.

Several options are: (a) the framework of SDF is well suited for the task; (b) ALS is a specific type of optimization algorithm that can be used. The updating equations can be explicitly derived in analogy with the derivation in *Subsection IV.A*. Note that the GEVD based approach can be used to efficiently initialize these optimization based methods.

Now that the overdetermined DC-CPD has been computed, we obtain $\tilde{\boldsymbol{C}}^{(m,n)} = \boldsymbol{C}^{(m,n)} \boldsymbol{D}_C^{(m,n)} \boldsymbol{\Pi} = \boldsymbol{B}^{(m,n)-T}$, and $\tilde{\boldsymbol{A}}^{(m)} \odot \tilde{\boldsymbol{A}}^{(n)*} = \boldsymbol{T}_3^{(m,n)} \boldsymbol{B}^{(m,n)}$, where $\tilde{\boldsymbol{A}}^{(m)} = \boldsymbol{A}^{(m)} \boldsymbol{D}_A^{(m)} \boldsymbol{\Pi}$ and $\tilde{\boldsymbol{A}}^{(n)} = \boldsymbol{A}^{(n)} \boldsymbol{D}_A^{(n)} \boldsymbol{\Pi}$ are estimates of $\boldsymbol{A}^{(m)}$ and $\boldsymbol{A}^{(n)}$ up to scaling and permutation ambiguities, $m, n = 1,\ldots,M$. Matrices $\boldsymbol{D}_A^{(m)}$ and $\boldsymbol{D}_A^{(n)}$ are diagonal matrices, and $\boldsymbol{D}_A^{(m)} \boldsymbol{D}_A^{(n)*} \boldsymbol{D}_C^{(m,n)} = \boldsymbol{I}_R$. We define $\boldsymbol{G}_r^{(m,n)} \triangleq \text{unvec}(\boldsymbol{T}_3^{(m,n)} \boldsymbol{B}^{(m,n)})_{(:,r)}$, and collect these matrices in an $NM \times NM$ matrix $\boldsymbol{G}_r$ as follows:

$$\boldsymbol{G}_r = \begin{bmatrix} \boldsymbol{G}_r^{(1,1)} & \cdots & \boldsymbol{G}_r^{(1,M)} \\ \vdots & \ddots & \vdots \\ \boldsymbol{G}_r^{(M,1)} & \cdots & \boldsymbol{G}_r^{(M,M)} \end{bmatrix} = \begin{bmatrix} \tilde{\boldsymbol{a}}_r^{(1)} \\ \vdots \\ \tilde{\boldsymbol{a}}_r^{(M)} \end{bmatrix} \cdot [\tilde{\boldsymbol{a}}_r^{(1)H},\ldots,\tilde{\boldsymbol{a}}_r^{(M)H}], \quad (20)$$

where $\tilde{\boldsymbol{a}}_r^{(m)}$ denotes the $r$th column of $\tilde{\boldsymbol{A}}^{(m)}$, $r = 1,\ldots,R$, $m = 1,\ldots,M$. We can calculate $[\tilde{\boldsymbol{a}}_r^{(1)T},\ldots,\tilde{\boldsymbol{a}}_r^{(M)T}]^T$ as the dominant eigenvector of $\boldsymbol{G}_r$.

We summarize the algebraic DC-CPD algorithm in Table I.

TABLE I
ALGEBRAIC DC-CPD ALGORITHM

| **Input:** $\{\mathcal{T}^{(m,n)} \in \mathbb{C}^{N \times N \times R} | 1 \le m,n \le M\}$ admitting DC-CPD (2). |
|---|
| **1:** Group $\mathcal{T}^{(m,n)}$ into $M$ sets $\Upsilon^{(m)}$ by (8). |
| **2:** For each pair of tensors $\mathcal{T}^{(m,g)}, \mathcal{T}^{(m,h)} \in \Upsilon^{(m)}$, perform coupled rank-1 detection mapping (14), for all frontal slices $\boldsymbol{T}_{(:,:,s)}^{(m,g)}$, $\boldsymbol{T}_{(:,:,u)}^{(m,h)}$, to obtain tensors $\Phi_{(1,2)}(\boldsymbol{T}_{(:,:,s)}^{(m,g)}, \boldsymbol{T}_{(:,:,u)}^{(m,h)})$, $1 \le s, u \le R$. Vectorize these tensors and stack them into the columns of matrix $\boldsymbol{\Gamma}^{(m,g,h)}$. |
| **3:** Calculate a set of $R$ basis vectors in $\ker(\boldsymbol{\Gamma}^{(m,g,h)})$, and reshape these vectors into matrices $\boldsymbol{V}_l^{(m,g,h)}, l = 1,\ldots,R$. Stack $\boldsymbol{V}_l^{(m,g,h)}$ with varying $l$ into a tensor $\mathcal{W}^{(m,g,h)}$. For varying $m, g, h$, the tensors $\mathcal{W}^{(m,g,h)}$ together admit an overdetermined DC-CPD (17) |
| **4:** Solve (17) to compute the matrices $\boldsymbol{B}^{(m,n)}$. Then we have $\tilde{\boldsymbol{C}}^{(m,n)} = \boldsymbol{B}^{(m,n)-T}$ and $\tilde{\boldsymbol{A}}^{(m)} \odot \tilde{\boldsymbol{A}}^{(n)*} = \boldsymbol{T}_3^{(m,n)} \boldsymbol{B}^{(m,n)}$. |
| **5:** Compute matrix $\boldsymbol{G}_r$ by (20). Then the $r$th column of the factor matrices $\tilde{\boldsymbol{A}}^{(m)}$ can be computed as the dominant eigenvector of $\boldsymbol{G}_r$, $m = 1,\ldots,M$, $r = 1,\ldots,R$. |
| **Output:** Estimates of the factor matrices $\tilde{\boldsymbol{A}}^{(m)}$ and $\tilde{\boldsymbol{C}}^{(m,n)}, 1 \le m,n \le M$. |

## IV. OPTIMIZATION BASED ALGORITHMS

In this section, we introduce two optimization based DC-CPD methods: (*i*) DC-CPD based on ALS, and (*ii*) DC-CPD based on SDF.

We note that in the overdetermined case, the DC-CPD can be viewed as the generalized joint diagonalization (GJD) of the frontal slices of the data tensors. Therefore, iterative GJD algorithms [36]–[39] can also be used to compute an overdetermined DC-CPD.

### A. DC-CPD based on ALS

The basic idea of ALS is to update each unknown factor in the LS sense with the other factor matrices fixed and alternate over such conditional updates. ALS monotonically decreases the cost function (4), but it is not guaranteed to converge to a stationary point. In cases where ALS converges, it does so at most at a linear rate near a (local) optimum [66], [67].

We make the following assumption, for $1 \le m, n \le M$, to

ensure that in each ALS step, the factor matrix update is unique:

$$\begin{bmatrix} \boldsymbol{A}^{(1)*} \odot \boldsymbol{C}^{(1,m)*} \\ \vdots \\ \boldsymbol{A}^{(m)*} \odot \boldsymbol{C}^{(m,m)*} \\ \boldsymbol{A}^{(m+1)*} \odot \boldsymbol{C}^{(m,m+1)} \\ \vdots \\ \boldsymbol{A}^{(M)*} \odot \boldsymbol{C}^{(m,M)} \end{bmatrix} \text{ and } \boldsymbol{A}^{(m)} \odot \boldsymbol{A}^{(n)*} \text{ have full column rank.} \quad (21)$$

Note that (21) is a necessary condition for the uniqueness of DC-CPD. In fact, if (21) does not hold, DC-CPD with a smaller number of terms is possible. More precisely, if $[(\boldsymbol{A}^{(1)*} \odot \boldsymbol{C}^{(1,m)*})^T, ..., (\boldsymbol{A}^{(m)*} \odot \boldsymbol{C}^{(m,m)*})^T, (\boldsymbol{A}^{(m+1)*} \odot \boldsymbol{C}^{(m,m+1)})^T, ..., (\boldsymbol{A}^{(M)*} \odot \boldsymbol{C}^{(m,M)})^T]^T$ does not have full column rank, e.g., if the following holds for fixed $m$,

$$\begin{bmatrix} \boldsymbol{a}_R^{(1)*} \otimes \boldsymbol{c}_R^{(1,m)*} \\ \vdots \\ \boldsymbol{a}_R^{(m)*} \otimes \boldsymbol{c}_R^{(m,m)*} \\ \boldsymbol{a}_R^{(m+1)*} \otimes \boldsymbol{c}_R^{(m,m+1)} \\ \vdots \\ \boldsymbol{a}_R^{(M)*} \otimes \boldsymbol{c}_R^{(m,M)} \end{bmatrix} = \sum_{r=1}^{R-1} \lambda_r^{(m)} \begin{bmatrix} \boldsymbol{a}_r^{(1)*} \otimes \boldsymbol{c}_r^{(1,m)*} \\ \vdots \\ \boldsymbol{a}_r^{(m)*} \otimes \boldsymbol{c}_r^{(m,m)*} \\ \boldsymbol{a}_r^{(m+1)*} \otimes \boldsymbol{c}_r^{(m,m+1)} \\ \vdots \\ \boldsymbol{a}_r^{(M)*} \otimes \boldsymbol{c}_r^{(m,M)} \end{bmatrix}, \quad (22)$$

then $\mathcal{T}^{(m,n)} = \sum_{r=1}^{R-1} (\boldsymbol{a}_r^{(m)} + \lambda_r^{(m)} \boldsymbol{a}_R^{(m)}) \otimes \boldsymbol{a}_r^{(n)*} \otimes \boldsymbol{c}_r^{(m,n)}$, and $\mathcal{T}^{(m,n')} = \sum_{r=1}^{R-1} \boldsymbol{a}_r^{(n')} \otimes (\boldsymbol{a}_r^{(m)} + \lambda_r^{(m)} \boldsymbol{a}_R^{(m)})^* \otimes \boldsymbol{c}_r^{(m,n')}$ for $n'=1,...,m$ and $n=m,...,M$. From a similar reasoning it follows that DC-CPD is possible with a smaller number of terms if $\boldsymbol{A}^{(m)} \odot \boldsymbol{A}^{(n)*}$ does not have full column rank.

We define $\mathcal{E}^{(m,n)} \triangleq \mathcal{T}^{(m,n)} - [\![\boldsymbol{A}^{(m)}, \boldsymbol{A}^{(n)*}, \boldsymbol{C}^{(m,n)}]\!]_R$ as the residue for $\mathcal{T}^{(m,n)}$, and $\eta_m \triangleq \sum_{n=m}^M \|\mathcal{E}^{(m,n)}\|_F^2 + \sum_{n'=1}^m \|\mathcal{E}^{(n',m)}\|_F^2$ as the sum of LS errors for the tensors that are coupled by $\boldsymbol{A}^{(m)}$, for fixed $m$. Then we have $\eta = 0.5 \cdot \sum_{m=1}^M \eta_m$. We denote $\boldsymbol{E}_1^{(m,n)}$ and $\boldsymbol{E}_2^{(m,n)}$ as the mode-1 and mode-2 matrix representation of $\mathcal{E}^{(m,n)}$, respectively. Then $\eta_m$ can be written as:

$$\eta_m = \sum_{n'=1}^m \|\boldsymbol{E}_2^{(n',m)*}\|_F^2 + \sum_{n=m}^M \|\boldsymbol{E}_1^{(m,n)}\|_F^2$$
$$= \left\| \begin{bmatrix} \boldsymbol{T}_2^{(1,m)*} \\ \vdots \\ 2\boldsymbol{T}_2^{(m,m)*} \\ \boldsymbol{T}_1^{(m,m+1)} \\ \vdots \\ \boldsymbol{T}_1^{(m,M)} \end{bmatrix} - \begin{bmatrix} \boldsymbol{A}^{(1)*} \odot \boldsymbol{C}^{(1,m)*} \\ \vdots \\ 2\boldsymbol{A}^{(m)*} \odot \boldsymbol{C}^{(m,m)*} \\ \boldsymbol{A}^{(m+1)*} \odot \boldsymbol{C}^{(m,m+1)} \\ \vdots \\ \boldsymbol{A}^{(M)*} \odot \boldsymbol{C}^{(m,M)} \end{bmatrix} \boldsymbol{A}^{(m)T} \right\|_F^2. \quad (23)$$

In (23) we have used the property that $\boldsymbol{T}_1^{(m,m)} = \boldsymbol{T}_2^{(m,m)*}$ and $\boldsymbol{C}^{(m,m)} = \boldsymbol{C}^{(m,m)*}$, due to the conjugated symmetry.

We update $\boldsymbol{A}^{(m)}$ by minimizing $\eta_m$ with $\boldsymbol{A}^{(1)*} \odot \boldsymbol{C}^{(1,m)*}, ..., \boldsymbol{A}^{(m)*} \odot \boldsymbol{C}^{(m,m)*}, \boldsymbol{A}^{(m+1)*} \odot \boldsymbol{C}^{(m,m+1)}, ..., \boldsymbol{A}^{(M)*} \odot \boldsymbol{C}^{(m,M)}$ fixed, which is an LS sub-problem that is linear in factor matrix $\boldsymbol{A}^{(m)}$.

Next, we update $\boldsymbol{C}^{(m,n)}$ for each $(m,n)$ as follows, with $\boldsymbol{A}^{(m)}$ and $\boldsymbol{A}^{(n)}$ fixed:

$$\boldsymbol{C}^{(m,n)} = \arg\min_{\boldsymbol{C}^{(m,n)}} \left\| \boldsymbol{T}_3^{(m,n)} - (\boldsymbol{A}^{(m)} \odot \boldsymbol{A}^{(n)*}) \boldsymbol{C}^{(m,n)T} \right\|_F^2. \quad (24)$$

We perform the ALS steps for $\boldsymbol{A}^{(m)}$ and $\boldsymbol{C}^{(m,n)}$, $1 \leq m, n \leq M$, by minimizing (23) and (24) in an alternating manner until convergence. To accelerate the convergence, enhanced line search (ELS) and exact line search/plane search strategies can be developed analogously to the CPD case [68]–[70]. Table II summarizes the ALS based DC-CPD algorithm.

TABLE II
DC-CPD: ALS ALGORITHM

| |
|---|
| **Input:** $\{\mathcal{T}^{(m,n)} \in \mathbb{C}^{N \times N \times R} \mid 1 \leq m, n \leq M\}$ admitting DC-CPD (2) |
| **1:** Initialize either randomly or with the results from the algebraic DC-CPD algorithm. |
| **2:** Do the following until convergence: |
| - Update $\boldsymbol{A}^{(m)}$ by minimizing (23) with $\boldsymbol{A}^{(1)*} \odot \boldsymbol{C}^{(1,m)*},...,\boldsymbol{A}^{(m)*} \odot \boldsymbol{C}^{(m,m)*}, \boldsymbol{A}^{(m+1)*} \odot \boldsymbol{C}^{(m,m+1)},...,\boldsymbol{A}^{(M)*} \odot \boldsymbol{C}^{(m,M)}$ fixed, $1 \leq m \leq M$. |
| - Update $\boldsymbol{C}^{(m,n)}$ by (24) with $\boldsymbol{A}^{(m)}$ and $\boldsymbol{A}^{(n)}$ fixed, $1 \leq m, n \leq M$. |
| **Output:** Estimates of the factor matrices: $\tilde{\boldsymbol{A}}^{(m)}$ and $\tilde{\boldsymbol{C}}^{(m,n)}$, $1 \leq m, n \leq M$. |

### B. DC-CPD based on SDF

The framework of SDF, embedded in Tensorlab 3.0 [65], allows a rapid prototyping of the analysis of one or more coupled and/or structured tensor datasets, which may be complex, sparse, or incomplete. It offers multiple choices of decompositions, regularizations, and structures of factor matrices. Through a domain specific language, the users can flexibly and easily formulate their specific SDF models. Tensorlab 3.0 offers two classes of numerical optimization algorithms to solve SDF problems: quasi-Newton (QN) methods and Nonlinear Least Squares (NLS) methods, implemented in the 'sdf_minf.m' and 'sdf_nls.m' functions, respectively.

Therefore, we can use SDF to implement an optimization based DC-CPD algorithm. In this paper, we use both the QN and the NLS based solvers for the computation of the DC-CPD. We note that the NLS method is of low per-iteration cost and close to quadratic convergence near a (local) optimum [14]. We do not elaborate on QN, NLS and SDF here. Instead, we refer the readers to [14] and [64] for more details.

## V. DISCUSSION

In this section, we discuss several theoretical and practical aspects of DC-CPD.

### A. Generic uniqueness conditions

*Theorem 2* provides deterministic uniqueness conditions under which DC-CPD can be calculated algebraically. We can also obtain the generic value of the upper bound of $R$, denoted as $R_{\max}$. We call a property generic if it holds with probability one, when the parameters it involves are drawn from continuous probability densities. Generically, the matrix $\boldsymbol{C}^{(m,n)} \in \mathbb{C}^{R \times R}$ has full column rank. Hence, the generic version of the uniqueness conditions in *Theorem 2* depends only on $N$. We give the generic value of $R_{\max}$, for a number of different $N$, in Table II. For comparison, we also list the generic value of $R_{\max}$ for CPD [10]. Note that in Table III, the third dimension of the tensors is required to be not less than $R_{\max}$, and that the generic values of $R_{\max}$ in Table III apply to all $M \geq 2$.

The numerical values of $R_{\max}$ can be obtained using *Theorem 2* and *Corollary 10* in p.9 of [71], which together imply that, for fixed $N$ and $R$, DC-CPD is generically unique if we can find one example for which the decomposition is unique, and it suffices to try a random example. We give a detailed explanation in *Section III* of the *Supplementary materials*. Hence, one only needs to check if the matrix $\boldsymbol{\Phi}_{g,h}^{(m)}$ (15), which holds all the vectors $\psi(\boldsymbol{a}_t^{(m)}\boldsymbol{a}_t^{(g)H},\boldsymbol{a}_r^{(m)}\boldsymbol{a}_r^{(h)H}), 1 \leq t \neq r \leq R$, has full column rank for randomly generated factor matrices $\boldsymbol{A}^{(1)},...,\boldsymbol{A}^{(M)}$. $R_{\max}$ is then chosen as the largest $R$ for which the matrices $\boldsymbol{\Phi}_{g,h}^{(m)}$, for all $1 \leq g < h \leq M, 1 \leq m \leq M$, have full column rank. We note that the uniqueness conditions of DC-CPD are more relaxed than those of CPD, thanks to the coupling structure exploited in DC-CPD.

TABLE III
GENERIC VALUE OF $R_{\max}$ FOR DC-CPD AND CPD.

| $N$ | 2 | 3 | 4 | 5 | 6 | 7 | 8 |
|---|---|---|---|---|---|---|---|
| DC-CPD | 2 | 5 | 10 | 16 | 23 | 32 | 42 |
| CPD | 2 | 4 | 9 | 14 | 21 | 30 | 40 |

*B. Algebraic vs. optimization*

The algebraic DC-CPD approach is guaranteed to return the exact solution in the exact case. When noise is present, the conventional linear algebra operations, such as the calculation of the basis vectors of the null space of a matrix as its singular vectors, and the matrix best rank-1 approximation, are by themselves optimal in LS sense. The overall result, however, is not guaranteed to be optimal and is actually (mildly) suboptimal in practice.

On the other hand, an optimization based DC-CPD algorithm (e.g. the ALS or SDF based DC-CPD), which directly maximizes the fit to the data tensors, provides the optimal results in the LS sense, if it converges to the global minimum. However, we have observed that in J-BSS, especially in underdetermined J-BSS, the optimization based DC-CPD algorithms are very sensitive to initialization, and sometimes do not return the correct results even in the noiseless case. Hence, in practice, we can use the algebraic algorithm to efficiently initialize the optimization based algorithms.

*C. DC-CPD vs. CPD*

Obviously, in a DC-CPD we may ignore the double coupling and try to obtain estimates of the factor matrices from each individual tensor $\mathcal{T}^{(m,n)}$. Therefore, by dropping the double coupling, we can perform CPD to each tensor $\mathcal{T}^{(m,n)}$ separately to obtain the estimates of factor matrices $\boldsymbol{A}^{(1)},...,\boldsymbol{A}^{(M)}$. It is then natural to ask what benefits DC-CPD has over CPD. We briefly mention the following.

First, as will be shown later in the experiments, DC-CPD may provide better robustness to noise and model errors than CPD as it exploits more information and the way it is coupled. As in J-BSS the coupling comes from the similarity of the sources across different datasets, DC-CPD is expected to perform well in finding components that are consistently present in multi-set signals.

Second, DC-CPD components from distinct datasets are automatically aligned. In other words, DC-CPD avoids the permutation alignment / parameter pairing in a post-processing step, which can otherwise be difficult and time-consuming.

Third, DC-CPD can be unique when none of its constituting CPDs is unique, as is indicated by the generic uniqueness results in *Subsection V.A*.

*D. Complexity of algebraic DC-CPD*

We note that the complexity of the algebraic DC-CPD algorithm is $O(M^3N^4R^2 + M^3N^4R^4 + 5.5M^3R^6)$ flops. The memory requirements of the algorithm are $O(0.5M^3R^3 + 0.5M^3N^4R^2)$ complex numbers. Please refer to *Subsection IV.B* of the *Supplementary materials* for a detailed analysis of the complexity and memory requirements.

Further, we note that the algebraic DC-CPD admits an efficient implementation. Instead of explicitly constructing $\boldsymbol{\Gamma}^{(m,g,h)}$, we can calculate the $R^2 \times R^2$ Hermitian matrices $\boldsymbol{\Omega}^{(m,g,h)} \triangleq \boldsymbol{\Gamma}^{(m,g,h)H}\boldsymbol{\Gamma}^{(m,g,h)}$, taking advantage of the algebraic structure of $\boldsymbol{\Gamma}^{(m,g,h)}$. We note that $\ker(\boldsymbol{\Omega}^{(m,g,h)}) = \ker(\boldsymbol{\Gamma}^{(m,g,h)})$, while $\boldsymbol{\Omega}^{(m,g,h)}$ has smaller size and can be computed more efficiently than $\boldsymbol{\Gamma}^{(m,g,h)}$. We provide the details in *Subsection IV.A* of the *Supplementary materials*. Compared with the original version of algebraic DC-CPD, this implementation has lower complexity and memory requirements, namely $O(M^3R^6 + 0.5M^3N^2R^4)$ flops and $O(0.5M^3R^4 + M^2N^2R^2)$ complex numbers, respectively. Note that the efficient implementation exploits the full double coupling structure, i.e. we do not sacrifice part of the available information to gain in terms of speed.

Although the complexity of algebraic DC-CPD may seem high at first sight, it is actually quite reasonable. Recall from *Subsection III.B* and *Subsection III.C*, that in the algebraic underdetermined framework for CPD and DC-CPD, the problems are transformed into overdetermined problems via the computation of the null space of a matrix that has at least $O(R^2)$ columns. In numerical linear algebra, factorizations of $M \times N$ matrices typically have a complexity proportional to $O(MN^2)$ [72]. In this perspective, the complexity of the algebraic DC-CPD is indeed very reasonable.

Note that $N < R$ in the above expressions. When $N \geq R$, there is no need to use the coupled rank-1 detection mapping, and the complexity is equal to that of a GEVD based scheme, which is much lower.

VI. EXPERIMENTS

In this section we present experimental results to demonstrate the performance of DC-CPD, in comparison with standard CPD. In experiment A, we discuss high-accuracy computation of exact decompositions. In experiment B, we compare DC-CPD and other BSS or J-BSS algorithms in J-BSS of synthetic noisy multi-set signals. In experiment C, we apply DC-CPD in wideband array signal direction-of-arrival (DOA) estimation. In experiment B and C, we consider both overdetermined and underdetermined cases.

We use the following abbreviations:
- DC-CPD-ALG: algebraic DC-CPD algorithm.
- DC-CPD-ALS: DC-CPD via ALS.
- DC-CPD-SDF(QN): quasi-Newton DC-CPD method, implemented with the 'sdf_minf.m' function in SDF.
- DC-CPD-SDF(NLS): NLS based DC-CPD method, implemented with the 'sdf_nls.m' function in SDF.
- SOBIUM: second-order blind identification of underdetermined mixtures [11]. It is based on the CPD of each auto-covariance tensor.
- CPD-C: variant of SOBIUM based on the CPD of each

cross-covariance tensor. Note that this method may generate multiple estimates of the same factor matrix, e.g. there are $(M-1)$ estimates of $A^{(m)}$ if we perform CPD separately on $\mathcal{T}^{(1,m)},...,\mathcal{T}^{(1,m-1)},\mathcal{T}^{(1,m+1)},...,\mathcal{T}^{(1,M)}$. Here we perform CPD on tensor $\mathcal{T}^{(m,m+1)}$ to estimate $A^{(m)}$ for $m=1,...,M-1$. The matrix $A^{(M)}$ is obtained as the second factor matrix of the CPD of $\mathcal{T}^{(M-1,M)}$ up to a conjugation.
- MCCA: multi-set canonical covariance analysis [33].
- GOJD: generalized orthogonal joint diagonalization of second-order cross-covariance matrices between multiple prewhitened datasets [37].

In the implementation of DC-CPD-ALG, we use the SD based scheme to solve the overdetermined DC-CPD in (17). In the implementation of DC-CPD-ALS, DC-CPD-SDF(QN), and DC-CPD-SDF(NLS), we initialize either randomly or by the results of DC-CPD-ALG, depending on the application. The initialization details will be given for each example below. For the calculation of CPD in the implementation of SOBIUM and CPD-C, we use the NLS based CPD algorithm, implemented via the 'cpd_nls.m' function in Tensorlab 3.0. The tolerance on the relative function value and relative step size in the stopping criteria of 'cpd_nls' is set to TolFun = $10^{-12}$ and TolX = $10^{-8}$, respectively. We initialize the NLS based CPD algorithm with the results from the algebraic CPD algorithm [10]. In experiments B and C, the data tensors $\mathcal{T}^{(m,n)}$ in (6) are approximated by their finite-sample version:

$$\left(\tilde{\mathcal{T}}^{(m,n)}\right)_{(:,:,k)} = L^{-1} \cdot \left[\sum_{l=1}^{L}\left(X^{(m,k)}\right)_{(:,l)}\left(X^{(n,k)}\right)_{(:,l)}^{H}\right], \quad (25)$$

where $X^{(m,k)} \in \mathbb{C}^{N \times L}$ denotes the $k$th temporal frame of $x^{(m)}(t)$ in (6) with frame length $L$, overlapping with adjacent frames with overlap ratio $\alpha \in [0,1]$.

In experiment C, we use the benchmark codes 'stft.m' and 'istft.m' from [73] to perform the short-time Fourier transform (STFT) and the inverse STFT (ISTFT). All the experiments are performed on a workstation with following configuration, CPU: Intel Xeon E5-2640 v4 @ 2.4 GHz; Memory: 128GB; System: 64bit Windows 10; MATLAB R2016a.

*A. Results for exact decompositions*

The data tensors are directly generated by (7), where both the real and imaginary parts of each entry of the factor matrices $A^{(m)} \in \mathbb{C}^{N \times R}$ and $C^{(m,n)} \in \mathbb{C}^{R \times R}$ are randomly drawn from a standard normal distribution, and no noise is added. We fix the third dimension of all the data tensors to $R$ and let the number of mixing matrices $M = 3$. The mean relative error of the estimates of all the loading matrices is defined as:

$$\varepsilon = M^{-1}\sum_{m=1}^{M} \min_{\Pi^{(m)},D^{(m)}} \left( \left\| A^{(m)} - \Pi^{(m)}D^{(m)}\tilde{A}^{(m)} \right\|_F^2 \Big/ \left\| A^{(m)} \right\|_F^2 \right), (26)$$

where $\tilde{A}^{(m)}$ is the estimate of the true mixing matrix $A^{(m)}$, $\Pi^{(m)}$ is a permutation matrix and $D^{(m)}$ is a diagonal matrix. For DC-CPD-ALS, DC-CPD-SDF(NLS) and DC-CPD-SDF(QN), we try ten random initial values and select the one that gives the best fit after the first ten iterations to effectively initialize the algorithm. As there is no noise, we want to estimate the factor matrices with high precision. Therefore, we terminate the ALS iteration when $|\xi_{cur} - \xi_{prev}|/\xi_{prev} \leq 10^{-16}$, where $\xi_{cur}$ and $\xi_{prev}$ denote the LS error in the current and previous iteration, respectively. For DC-CPD-SDF(NLS) and DC-CPD-SDF(QN), we set the termination parameters 'TolFun' and 'TolX' in 'sdf_nls.m' and 'sdf_minf' to $10^{-16}$ and $10^{-8}$, respectively. In addition, we set the maximal number of iterations to 2000 for DC-CPD-ALS, DC-CPD-SDF(NLS) and DC-CPD-SDF(QN). The mean relative error and CPU time (denoted as '$t$') of the compared algorithms for different values of $N$ and $R$ are reported in Table IV. Note that, under the listed conditions, the rank $R$ is high and CPD has been proven to be generically not unique [18]. All the results in Table IV are averaged over 100 independent runs.

We see that in all the considered cases, DC-CPD-ALG has found the exact solution. DC-CPD-ALS, DC-CPD-SDF(NLS) and DC-CPD-SDF(QN), however, do not always find the correct solution in all cases, even in the absence of noise. This can be seen from the histograms of the mean relative errors of DC-CPD-ALS, DC-CPD-SDF(NLS) and DC-CPD-SDF(QN). By way of example, the histogram for setting $N = 3$, $R = 5$ is shown in Fig.2. Further, Table IV shows that DC-CPD-ALG provides more efficient computation than DC-CPD-ALS, DC-CPD-SDF(NLS) and DC-CPD-SDF(QN) in exact decompositions. In practice, noise is usually present, and the performance of DC-CPD-ALG will deteriorate, but can still be used as a low-cost initialization for the optimization based methods, which directly fit the data tensor in the LS sense. This will be shown later.

TABLE IV
PERFORMANCE OF DC-CPD-ALG, DC-CPD-ALS, DC-CPD-SDF(NLS), AND DC-CPD-SDF(QN) IN EXACT DECOMPOSITIONS, '$N$' IS THE NUMBER OF OBSERVATIONS, '$R$' IS THE NUMBER OF SOURCES, '$\varepsilon$' IS THE MEAN RELATIVE ERROR, AND '$t$' IS THE MEAN CPU TIME.

|  | DC-CPD-ALG | DC-CPD-ALS | DC-CPD-SDF (NLS) | DC-CPD-SDF (QN) |
|---|---|---|---|---|
| $N = 3$ | $\varepsilon = 7.05 \times 10^{-15}$ | $\varepsilon = 0.0906$ | $\varepsilon = 0.0065$ | $\varepsilon = 0.1899$ |
| $R = 5$ | $t = 0.059$ sec. | $t = 15.78$ sec. | $t = 22.37$ sec. | $t = 28.78$ sec. |
| $N = 4$ | $\varepsilon = 8.91 \times 10^{-13}$ | $\varepsilon = 0.2797$ | $\varepsilon = 0.2338$ | $\varepsilon = 0.5296$ |
| $R = 10$ | $t = 0.1241$ sec. | $t = 20.54$ sec. | $t = 219.04$ sec. | $t = 39.43$ sec. |
| $N = 5$ | $\varepsilon = 1.44 \times 10^{-12}$ | $\varepsilon = 0.4937$ | $\varepsilon = 0.4955$ | $\varepsilon = 0.6586$ |
| $R = 16$ | $t = 0.4609$ sec. | $t = 28.26$ sec. | $t = 314.5$ sec. | $t = 43.16$ sec. |
| $N = 6$ | $\varepsilon = 4.68 \times 10^{-13}$ | $\varepsilon = 0.5763$ | $\varepsilon = 0.6311$ | $\varepsilon = 0.7138$ |
| $R = 23$ | $t = 1.5765$ sec. | $t = 41.47$ sec. | $t = 375.69$ sec. | $t = 53.96$ sec. |
| $N = 7$ | $\varepsilon = 7.34 \times 10^{-13}$ | $\varepsilon = 0.6961$ | $\varepsilon = 0.7134$ | $\varepsilon = 0.7591$ |
| $R = 32$ | $t = 5.3349$ sec. | $t = 63.2$ sec. | $t = 457.56$ sec. | $t = 60.61$ sec. |

*B. J-BSS of multi-set signals*

In this experiment we use the finitely sampled version of the multi-set data model (5). The number of samples in each dataset is denoted as $Q$. We generate the $Q \times R$ source matrix $S^{(m)}$ from auxiliary matrix $S'_r$ of size $Q \times M$ as follows:

$$S_r^T = Q_r S_r'^T, \quad (27)$$

where $S_r \triangleq [(S^{(1)})_{:,r},...,(S^{(M)})_{:,r}] \in \mathbb{C}^{Q \times M}$, $(S^{(m)})_{:,r}$ denotes the

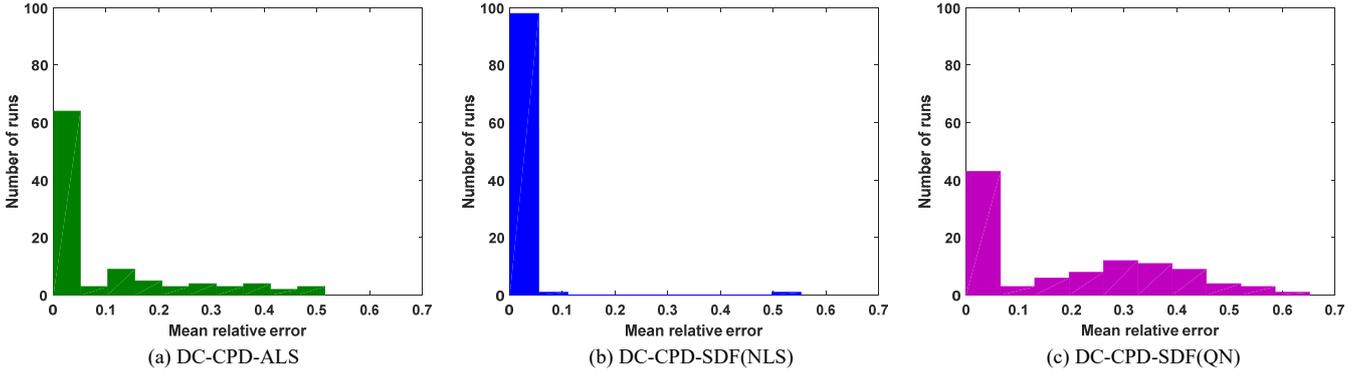

Fig.2. Histogram of mean relative error for DC-CPD-ALS, DC-CPD-SDF(NLS) and DC-CPD-SDF(QN), in *Experiment A*. $N = 3$, $R = 5$.

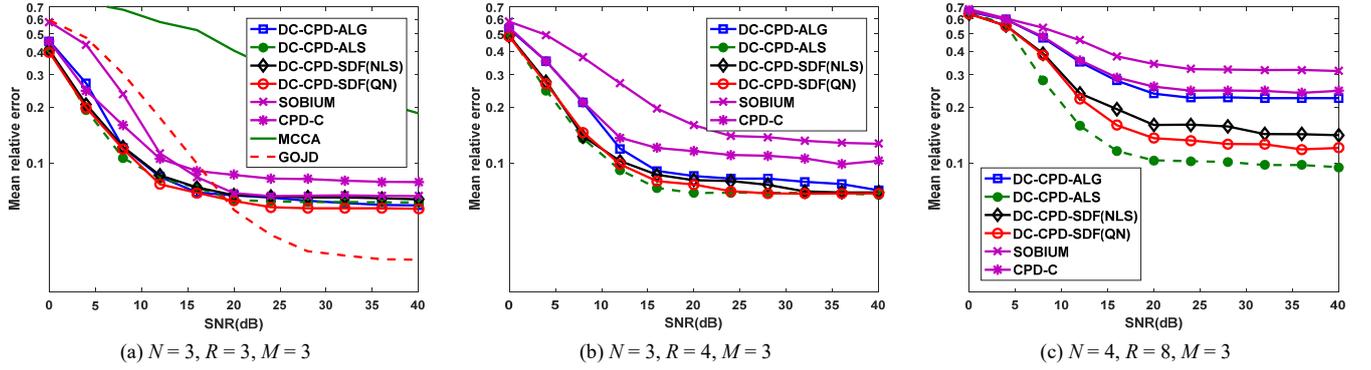

Fig.3. Mean relative error of DC-CPD-ALG, DC-CPD-ALS, DC-CPD-SDF(NLS), DC-CPD-SDF(QN), SOBIUM, and CPD-C vs. SNR in *Experiment B*. The plots illustrate the performance in (a) an overdetermined setting, (b) a slightly underdetermined setting, and (c) a highly underdetermined setting. In setting (a) MCCA and GOJD are also included in the comparison.

$r$th source vector in the $m$th dataset and $\boldsymbol{Q}_r \in \mathbb{C}^{M \times M}$ is a full rank matrix used to introduce inter-set dependence between the corresponding source signals in different datasets. Both the real and imaginary part of each entry of $\boldsymbol{Q}_r$ are drawn from a standard normal distribution. The underlying generating source matrix $\boldsymbol{S}'_r(t) \triangleq [\boldsymbol{s}'^{(1)}_r, ..., \boldsymbol{s}'^{(M)}_r]$ consists of complex binary phase shift keying (BPSK) signals that are amplitude modulated across $P$ time slots of length $L'$:

$$\boldsymbol{s}'^{(m)}_r = [\eta_1 \cdot \boldsymbol{s}'^{(m)T}_{r,1}, ..., \eta_P \cdot \boldsymbol{s}'^{(m)T}_{r,P}]^T \in \mathbb{C}^Q, \quad (28)$$

where $\eta_1, ..., \eta_P$ are amplitude modulation coefficients that are randomly drawn from a uniform distribution over [0, 1], and where $Q = PL'$. The sub-vector $\boldsymbol{s}'^{(m)}_{r,p}$ of length $L'$ is a complex BPSK sequence with entries chosen from symbols $\{1, -1\}$ with equal probability. By definition, $\boldsymbol{s}'^{(m)}_r$ is the concatenation of $P$ BPSK sequences, the amplitudes of which have been modulated by coefficients $\eta_1, ..., \eta_P$. In this experiment, $P$ is fixed to 20 and $L' = Q/20$.

The mixtures are constructed as:

$$\boldsymbol{X}^{(m)} = \sigma_s \frac{\boldsymbol{A}^{(m)} \boldsymbol{S}^{(m)T}}{\|\boldsymbol{A}^{(m)} \boldsymbol{S}^{(m)T}\|_F} + \sigma_n \frac{\boldsymbol{N}^{(m)}}{\|\boldsymbol{N}^{(m)}\|_F}, \quad m = 1, ..., M, \quad (29)$$

where the real and imaginary part of each entry of $\boldsymbol{A}^{(m)}$ are drawn from a standard normal distribution. Matrix $\boldsymbol{N}^{(m)}$ denotes the white Gaussian noise term added in the $m$th dataset, and $\sigma_s$, $\sigma_n$ denote the signal level and noise level, respectively. The signal-to-noise ratio (SNR) is defined as:

$$\text{SNR} = 20 \log_{10}(\sigma_s / \sigma_n). \quad (30)$$

DC-CPD-ALS, DC-CPD-SDF(NLS) and DC-CPD-SDF(QN) are initialized with the results of DC-CPD-ALG. We terminate the ALS iteration in DC-CPD-ALS when $|\xi_{cur} - \xi_{prev}|/\xi_{prev} \leq 10^{-7}$. For DC-CPD-SDF(NLS) and DC-CPD-SDF(QN), we set the termination parameters 'TolFun' and 'TolX' in 'sdf_nls.m' and 'sdf_minf.m' to $10^{-7}$ and $10^{-7}$, respectively.

We demonstrate the accuracy of the compared methods, in terms of mean relative error, in the following three settings: (a) an overdetermined case $N = 3$, $R = 3$, $M = 3$; (b) a slightly underdetermined case $N = 3$, $R = 4$, $M = 3$; and (c) a highly underdetermined case $N = 4$, $R = 8$, $M = 3$. In setting (a), we fix the framelength $L = 50$, the overlap ratio $\alpha = 0.5$, the number of frames $T = 39$. In setting (b), we set $L = 100$, $\alpha = 0.5$, $T = 39$. In setting (c), we set $L = 250$, $\alpha = 0.5$, $T = 39$. By definition, the number of samples is dependent on $L$ and $T$ via $Q = L(T + 1)/2$.

In all the three settings, DC-CPD-ALG, DC-CPD-ALS, DC-CPD-SDF(NLS), DC-CPD-SDF(QN), SOBIUM, and CPD-C are performed. MCCA and GOJD are performed only in setting (a) as they are algorithms for overdetermined J-BSS. We provide more results, with different parameters $L$, $T$, $M$, in *Subsection V.A* of the *Supplementary materials*. The results are in general consistent with those shown here, and they give additional insight in how the performance of different methods depends on these parameters.

We let SNR vary from 0dB to 40dB. The mean relative error $\varepsilon$ versus SNR is plotted in Fig.3. We perform 200 Monte Carlo runs to calculate each point in the curves.

In Fig.3, we can clearly see that all the DC-CPD algorithms perform better than both SOBIUM and CPD-C in all the three settings, as they use the information in both the auto-covariance and cross-covariance tensors, as well as the way the tensors are coupled. SOBIUM and CPD-C, on the other hand, use only auto-covariance or cross-covariance tensors. In setting (a), we see for all SNR levels, that the DC-CPD algorithms perform better than MCCA. They perform better than GOJD for low SNR (0dB–20dB), and slightly worse than the latter for high SNR (20dB–40dB). The good performance of GOJD at high SNR is mainly the result of prewhitening, which is possible only in the overdetermined case. In addition, we observe that DC-CPD-ALS, DC-CPD-SDF(NLS) and DC-CPD-SDF(QN) yield more accurate results than DC-CPD-ALG. This shows that the optimization based DC-CPD methods, which directly optimize the LS cost function, are more accurate than DC-CPD-ALG in the presence of noise and errors caused by finite sample effects. Overall, the difference is not very large. As expected, it is more significant when the problem is more challenging.

Comparing the results in Fig.3, including those in *Subsection V.A* of the *Supplementary materials*, we note that the J-BSS problem itself gets more sensitive to noise and model errors (e.g. finite sample effects) when it becomes more underdetermined. For example, in the overdetermined setting (a), all the DC-CPD algorithms perform well at medium and high SNR levels, even if the framelength $L$ is as small as 50. In the slightly underdetermined setting (b), DC-CPD has comparable accuracy as in setting (a), with the framelength $L$ increased to 100. In the highly underdetermined setting (c), the accuracy of DC-CPD is worse than in (a) and (b), even with $L = 250$.

We also provide results on the mean CPU time versus SNR and $M$, in *Subsection V.B* of the *Supplementary materials*. The results generally show that DC-CPD-ALG is faster than the optimization based DC-CPD methods and the NLS based CPD methods, SOBIUM and CPD-C. In the overdetermined case, DC-CPD-ALG is also faster than GOJD. The above results hold consistently for varying $M$.

*C. Wideband DOA estimation via a small-scaled array*

Wideband DOA estimation is a challenging problem in array processing as the conventional methods based on linear instantaneous mixing model only apply to narrowband signals. The conventional way to handle wideband signals is to transform them into multiple narrowband signals in different frequency bins, use narrowband techniques for each bin separately, and integrate these results (see [74], [75] and references therein). In this example, we will illustrate how J-BSS can be used to fuse the signals in adjacent frequency bins to improve the performance of wideband DOA estimation. In this experiment, we will use a small-scaled array, which is of particular interest for space-limited applications such as air-borne array signal processing. We mainly consider the far-field mixing model in this experiment. Note that our method does not impose any specific assumption on the structure of the mixing matrix, and thus also applies to other models. The two array configurations used in this experiment are shown in Fig.4.

The wideband output signal $y_n(t)$ of the $n$th sensor is given by:

$$y_n(t) = \sum_r s_r(t - \tau_{n,r}) + \varepsilon_n(t), \quad (31)$$

where $s_r(t)$ and $\varepsilon_n(t)$ denote the $r$th source signal and the $n$th noise term at time instance $t$, respectively. We denote $\mathbf{y}(t) \triangleq [y_1(t),...,y_N(t)]^T$ and $\boldsymbol{\varepsilon}(t) \triangleq [\varepsilon_1(t),...,\varepsilon_N(t)]^T$. The parameter $\tau_{n,r}$ denotes the time delay of the $r$th source from the reference point (e.g. the origin of the Cartesian coordinate system) to the $n$th sensor. It is given by:

$$\tau_{n,r} = \mathbf{k}_n^T \mathbf{p}_{(\theta_r, \varphi_r)} / c, \quad (32)$$

where $\mathbf{p}_{(\theta,\varphi)} \triangleq [\cos\theta\sin\varphi, \sin\theta\sin\varphi, \cos\varphi]^T$ is the direction vector with azimuth $\theta$ and elevation $\varphi$ (see Fig.4. for the definition of the angles), and $\theta_r$ and $\varphi_r$ are the azimuth and elevation of the $r$th source signal, respectively. The vector $\mathbf{k}_n \in \mathbb{R}^3$ holds the Cartesian coordinates of the $n$th sensor, and $c$ denotes the wave propagation speed.

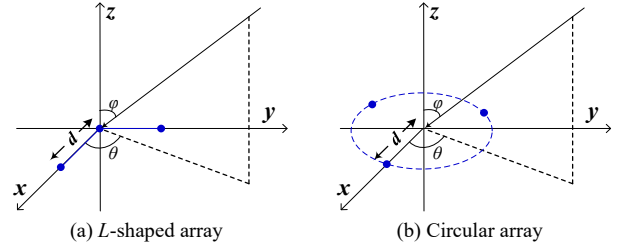

(a) L-shaped array     (b) Circular array

Fig.4. Array configurations and angle definition in experiment C.

We assume that there are $Q$ time samples. Due to finite sampling, the time delay $\tau_{n,r}$ is usually equal to a fractional number of time samples, which makes it difficult to construct the array signals directly by (31) in simulations. Here we use the scheme in [74] to generate the array signals. More in detail, according to [74], if the source signals have finite bandwidth and the observation time is much longer than the time delay, then the time delay corresponds to the phase shift of the Fourier coefficients in the Fourier domain. Therefore, we can construct the time delayed source signals by phase shifting their Fourier coefficients.

Now we explain the construction of the multi-set data for J-BSS. We convert the array output signal to the time-frequency domain via STFT with frame length $F$. By definition, there are $F$ frequency bins. If the bandwidth of each bin, $f_s / F$, is small enough, the array signal in each frequency bin can be approximately written as:

$$\mathbf{Y}^{(f)} = \sum_r \mathbf{a}_r^{(f)} \mathbf{s}_r^{(f)T} + \mathbf{N}^{(f)}, \quad f = 1,...,F, \quad (33)$$

where $\mathbf{Y}^{(f)}, \mathbf{N}^{(f)} \in \mathbb{C}^{N \times Q'}$ and $\mathbf{s}_r^{(f)} \in \mathbb{C}^{Q'}$ are the narrowband component of the array output, the noise term, and the $r$th source signal in the $f$th frequency bin, respectively. Parameter $Q'$ denotes the number of time samples of array signal $\mathbf{Y}^{(f)}$ in each frequency bin. Vector $\mathbf{a}_r^{(f)} \in \mathbb{C}^N$ is the steering vector of the $r$th source signal in the $f$th frequency bin:

$$\mathbf{a}_r^{(f)} \triangleq \exp(-i2\pi c^{-1} f) \cdot \mathbf{K}^T \mathbf{p}_{(\theta_r, \varphi_r)}, \quad (34)$$

where $\boldsymbol{K} = [\boldsymbol{k}_1,...,\boldsymbol{k}_N] \in \mathbb{R}^{3 \times N}$.

We select the $M$ successive frequency bins of which the array signals account for most of the energy as the multi-set data for J-BSS. After J-BSS is performed to estimate the associated steering vectors for the selected frequency bins, the DOAs are estimated by solving (34).

In this experiment, the source signal $\boldsymbol{s}_r$ is generated as: $\boldsymbol{s}_r = \boldsymbol{s}'_r * \boldsymbol{s}_c \in \mathbb{R}^Q$, where $\boldsymbol{s}_c$ is a single-tone carrier signal with frequency $f_c = 100\,\text{MHz}$. The baseband signal $\boldsymbol{s}'_r$ is constructed as in (28): $\boldsymbol{s}'_r = [\eta_1 \cdot \boldsymbol{s}'^T_{r,1},...,\eta_P \cdot \boldsymbol{s}'^T_{r,P}]^T \in \mathbb{R}^Q$, where $\eta_1,...,\eta_P$ are amplitude modulation coefficients that are randomly drawn from a uniform distribution over [0, 1]. The sub-vector $\boldsymbol{s}'_{r,p}$ of length $L'$ is a real-valued BPSK sequence with symbols chosen from $\{1, -1\}$ with equal probability. Here $P$ is fixed to 16 and thus $L' = Q/16$. For the construction of $\boldsymbol{s}'_{r,p}$, we let the bitrate $f_b = 50\,\text{MHz}$ and the sampling rate $f_s = 2500\,\text{MHz}$. Note that the source signals have a relative bandwidth of $f_b / f_c = 0.5$. The array mixture is generated using the scheme in [74], which well satisfied the pure delay model (31). The noise term is generated as a white Gaussian signal.

In the construction of the multi-set data, the framelength for STFT is fixed to $F = 256$. By definition, the bandwidth of each frequency bin is $f_s / F = 9.77\,\text{MHz} = 0.0977 f_c$, and thus the array signal in each frequency bin admits the narrowband formulation (33). Note that the number of time samples in each frequency bin is $Q' = 2Q/F = Q/128$.

We select the signals from the eighth to the fifteenth frequency bin as the multi-set data for J-BSS. Therefore, the number of datasets is $M = 8$. We calculate by (25), for a chosen framelength $L = Q'/16$ and overlap ratio $\alpha = 0.5$, the second-order covariance tensors as the data tensors for DC-CPD. We initialize DC-CPD-ALS, DC-CPD-SDF(NLS), and DC-CPD-SDF(QN) by the results of DC-CPD-ALG. The ALS iteration in DC-CPD-ALS is terminated when $|\xi_{cur} - \xi_{prev}|/\xi_{prev} \le 10^{-7}$. For DC-CPD-SDF(NLS) and DC-CPD-SDF(QN), we set the termination parameters 'TolFun' and 'TolX' in 'sdf_nls.m' and 'sdf_minf' to $10^{-7}$ and $10^{-7}$, respectively. Note that the accuracy of DOA estimation depends on how well the steering vectors are identified. We use the mean relative estimation error (26) to evaluate the performance.

We consider the following two cases: (a) an overdetermined setting: three sources impinge on an $L$-shaped array of interspacing $d$, as shown in Fig.4 (a), from DOAs $(\theta_1, \varphi_1) = (30°, 15°)$, $(\theta_2, \varphi_2) = (90°, 45°)$, and $(\theta_3, \varphi_3) = (150°, 75°)$; (b) a highly underdetermined setting: five sources impinge on a circular array of radius $d$, as shown in Fig.5 (b), from DOAs $(\theta_1, \varphi_1) = (18°, 9°)$, $(\theta_2, \varphi_2) = (54°, 27°)$, $(\theta_3, \varphi_3) = (90°, 45°)$, $(\theta_4, \varphi_4) = (126°, 63°)$, and $(\theta_5, \varphi_5) = (162°, 81°)$. In both cases, the number of time samples in each frequency bin is set to $Q' = 400$, and $d$ is set to half the wavelength of the signal in the highest frequency bin. We include DC-CPD-ALG, DC-CPD-ALS, DC-CPD-SDF(NLS), DC-CPD-SDF(QN), SOBIUM, and CPD-C in the comparison. In the first setting, we also include MCCA and GOJD in the comparison.

The SNR is defined in (30) with $\sigma_s$ and $\sigma_n$ denoting the signal and noise levels in the time domain, respectively. In the experiment, we let SNR vary from 0dB to 20dB. The mean relative error of the compared algorithms in the above two cases is plotted in Fig.5.

In Fig.5 (a) we observe that all the DC-CPD algorithms provide quite accurate results in the overdetermined setting. DC-CPD-ALG and DC-CPD-ALS perform about the same. DC-CPD-SDF(NLS) and DC-CPD-SDF(QN) are slightly more accurate. Comparing with CPD based methods, SOBIUM and CPD-C, we note that the DC-CPD algorithms are significantly more accurate at low SNR, while slightly less accurate at high SNR. For GOJD, we observe that it has good performance at high SNR, and poor performance at low SNR. The good performance of GOJD at high SNR is mainly the result of pre-whitening. We also see that all the DC-CPD and CPD based methods are more accurate than MCCA.

In Fig.5 (b) we see that SOBIUM and CPD-C do not provide correct result in this highly underdetermined setting. On the other hand, the DC-CPD algorithms still perform quite well. DC-CPD-ALS, DC-CPD-SDF(NLS) and DC-CPD-SDF(QN) perform slightly better than DC-CPD-ALG.

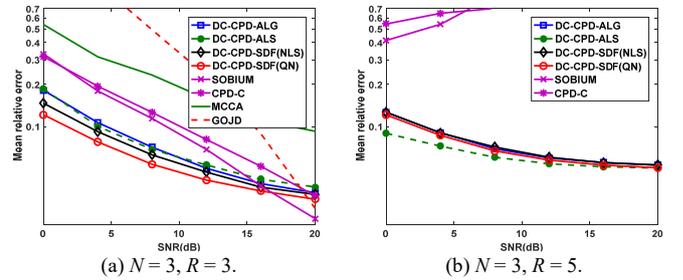

Fig.5. Mean relative error of DC-CPD-ALG, DC-CPD-ALS, DC-CPD-SDF(NLS), DC-CPD-SDF(QN), SOBIUM, CPD-C in wideband DOA estimation. The plots illustrate the performance in (a) an overdetermined setting, and (b) a highly underdetermined setting. In setting (a), MCCA and GOJD are also included in the comparison.

## VII. CONCLUSION

In this study, we have considered the problem of DC-CPD for J-BSS of multi-set signals. We have shown how, via the use of second-order statistics, a DC-CPD problem can be obtained from the multi-set signals in J-BSS. We have proposed an algebraic DC-CPD algorithm, via the use of the coupled rank-1 detection mapping, which is of particular interest in underdetermined problems. Deterministic and generic uniqueness conditions of DC-CPD have been studied, and shown to be more relaxed than their CPD equivalents. The proposed algorithm is deterministic and returns the exact solution in the noiseless case, as long as the uniqueness conditions are satisfied. When noise is present, the proposed algorithm can serve to provide a low-cost initialization for optimization based algorithms. We have also introduced optimization based DC-CPD methods, based on ALS and SDF, respectively. Experiments are performed to compare DC-CPD based J-BSS methods, CPD based BSS methods and several existing J-BSS methods, in both overdetermined and underdetermined J-BSS problems. The results have generally shown the superiority of using DC-CPD for J-BSS, with respect to uniqueness and accuracy.

# APPENDIX I
## PROOF OF THEOREM 3

We prove the following partial statements:

(i) $\ker(\boldsymbol{\Gamma}^{(m,g,h)}) = \ker(\boldsymbol{P}_{g,h}^{(m)T})$.

(ii) $\dim(\boldsymbol{\Gamma}^{(m,g,h)}) = R$.

(iii) The basis vectors $\boldsymbol{w}_r^{(m,g,h)}$ in $\ker(\boldsymbol{\Gamma}^{(m,g,h)})$, $r = 1,...,R$, can be written as linear combination of vectors $\boldsymbol{b}_u^{(m,g)} \otimes \boldsymbol{b}_u^{(m,h)}$, $u=1,...,R$, where $[\boldsymbol{b}_1^{(m,l)},...,\boldsymbol{b}_R^{(m,l)}] = \boldsymbol{B}^{(m,l)} \triangleq (\boldsymbol{C}^{(m,l)} \boldsymbol{D}_C^{(m,l)} \boldsymbol{\Pi})^{-T} \in \mathbb{C}^{R \times R}$, $\boldsymbol{D}_C^{(m,l)}$ is a diagonal matrix, and $\boldsymbol{\Pi}$ is a permutation matrix that is common to all the matrices $\boldsymbol{C}^{(m,l)}$, $l \in \{g,h\}$.

Note that $m$, $g$, $h$ are fixed in the above statements.

First, as $\boldsymbol{\Phi}_{g,h}^{(m)}$ is assumed to have full column rank, and $\boldsymbol{\Gamma}^{(m,g,h)} = \boldsymbol{\Phi}_{g,h}^{(m)} \cdot \boldsymbol{P}_{g,h}^{(m)T}$, we have $\ker(\boldsymbol{\Gamma}^{(m,g,h)}) = \ker(\boldsymbol{P}_{g,h}^{(m)T})$.

Second, as the matrices $\boldsymbol{C}^{(m,g)}, \boldsymbol{C}^{(m,h)}$ have full column rank, the Kronecker product $\boldsymbol{C}^{(m,g)} \otimes \boldsymbol{C}^{(m,h)}$ also has full column rank. The $R^2 \times (R^2 - R)$ matrix $\boldsymbol{P}_{g,h}^{(m)}$ has full column rank as well, since its columns constitute a subset of the columns of $\boldsymbol{C}^{(m,g)} \otimes \boldsymbol{C}^{(m,h)}$. By the rank nullity theorem, which states that $\dim[\ker(\boldsymbol{A})] + \text{rank}(\boldsymbol{A}) = J$ for any matrix $\boldsymbol{A}$ of size $I \times J$, we have:

$$\dim[\ker(\boldsymbol{P}_{g,h}^{(m)T})] = R^2 - \text{rank}(\boldsymbol{P}_{g,h}^{(m)T}) = R.$$

Third, we prove that $\boldsymbol{b}_r^{(m,g)} \otimes \boldsymbol{b}_r^{(m,h)}$ is in $\ker(\boldsymbol{P}_{g,h}^{(m)T})$, $r = 1,...,R$. This is equivalent to proving the following for all the values $1 \leq s, u \leq R$:

$$\begin{aligned} \left(\boldsymbol{c}_s^{(m,g)} \otimes \boldsymbol{c}_u^{(m,h)}\right)^T &\left(\boldsymbol{b}_r^{(m,g)} \otimes \boldsymbol{b}_r^{(m,h)}\right) \\ &= \left(\boldsymbol{c}_s^{(m,g)T} \boldsymbol{b}_r^{(m,g)}\right) \otimes \left(\boldsymbol{c}_u^{(m,h)T} \boldsymbol{b}_r^{(m,h)}\right) = 0. \end{aligned} \quad (35)$$

Recall that $\boldsymbol{b}_r^{(m,l)}$ is the $r$th column of $\tilde{\boldsymbol{C}}^{(m,l)-T}$, where $\tilde{\boldsymbol{C}}^{(m,l)} = \boldsymbol{C}^{(m,l)} \boldsymbol{D}_C^{(m,l)} \boldsymbol{\Pi}$. We assume the matrix $\boldsymbol{\Pi}$ permutes the columns of $\boldsymbol{C}^{(m,l)}$ in such a way that $(\boldsymbol{C}^{(m,l)} \boldsymbol{\Pi})_{:,r} = \boldsymbol{c}_k^{(m,l)}$, where $l \in \{g,h\}, r = \pi(k), 1 \leq k \leq R$, and $(\pi(1),...,\pi(R))$ is a permutation of $(1,...,R)$. We denote $\tilde{\boldsymbol{c}}_r^{(m,l)}$ as the $r$th column of $\tilde{\boldsymbol{C}}^{(m,l)}$. By definition we have $\tilde{\boldsymbol{c}}_r^{(m,l)} = d_r^{(m,l)} \boldsymbol{c}_k^{(m,l)}$, where $d_r^{(m,l)}$ is the $r$th entry on the diagonal of $\boldsymbol{D}_C^{(m,l)}$.

Therefore, we have $\boldsymbol{c}_t^{(m,l)T} \boldsymbol{b}_r^{(m,l)} = 0$ if $r = \pi(k)$ and $k \neq t$, with $l \in \{g,h\}, t \in \{s,u\}$. According to the definition of $\boldsymbol{P}_{g,h}^{(m)}$, we have $s \neq u$. Therefore, either $\boldsymbol{c}_s^{(m,g)T} \boldsymbol{b}_r^{(m,g)}$ or $\boldsymbol{c}_u^{(m,h)T} \boldsymbol{b}_r^{(m,h)}$ is equal to zero, which proves (35). Moreover, as $\boldsymbol{b}_1^{(m,l)},...,\boldsymbol{b}_R^{(m,l)}$ are linearly independent, the set $\{\boldsymbol{b}_r^{(m,g)} \otimes \boldsymbol{b}_r^{(m,h)}, r = 1,...,R\}$ contains $R$ linearly independent vectors, and thus is a maximal set of linearly independent vectors of $\ker(\boldsymbol{\Gamma}^{(m,g,h)})$. Therefore, the basis vectors $\boldsymbol{w}_r^{(m,g,h)}$ of $\ker(\boldsymbol{\Gamma}^{(m,g,h)})$ can be expressed as a linear combination of $\boldsymbol{b}_r^{(m,g)} \otimes \boldsymbol{b}_r^{(m,h)}, r = 1,...,R$. ∎

## ACKNOWLEDGMENT

The authors would like to express their sincere gratitude to Dr. Ignat Domanov and Dr. Mikael Sorensen for their insightful comments on the algebraic DC-CPD algorithm.

# Double Coupled Canonical Polyadic Decomposition for Joint Blind Source Separation

Xiao-Feng Gong, *Member, IEEE*, Qiu-Hua Lin, *Member, IEEE*, Feng-Yu Cong, *Senior Member, IEEE*, and Lieven De Lathauwer, *Fellow, IEEE*

Supplementary materials

## I. Construction of a Conjugated Symmetric DC-CPD by Tensor Concatenation

In the main manuscript, we assume that the DC-CPD has a conjugated symmetric structure: $\mathcal{T}^{(m,n)} = \text{perm}_{(2,1,3)}(\mathcal{T}^{(n,m)*})$, where $\text{perm}_{(2,1,3)}(\mathcal{T}^{(n,m)*})$ permutes the first and second indices of $\mathcal{T}^{(n,m)*}$. Here we explain how to create this symmetry, in cases where it is not readily present, via tensor concatenation.

The tensor concatenation operator "$\text{cat}(\cdot)$" constructs a tensor $\mathcal{T} \in \mathbb{C}^{I \times J \times (K+L)}$ by concatenation of two tensors $\mathcal{X} \in \mathbb{C}^{I \times J \times K}$, $\mathcal{Y} \in \mathbb{C}^{I \times J \times L}$ along the third mode, as follows:

$$(\mathcal{T})_{i,j,k} \triangleq \begin{cases} (\mathcal{X})_{i,j,k}, & 1 \leq k \leq K, \\ (\mathcal{Y})_{i,j,k-K}, & K < k \leq K+L. \end{cases} \quad (S1)$$

Let us denote the tensors in a non-symmetric DC-CPD as $\mathcal{T}'^{(m,n)} = [\![A^{(m)}, A^{(n)*}, C'^{(m,n)}]\!]_R$, and group these tensors into $M$ sets $\Upsilon'^{(1)}, \ldots, \Upsilon'^{(M)}$. Each set $\Upsilon'^{(m)}$ contains $2M$ tensors that have a common factor matrix $A^{(m)}$, for fixed $m$:

$$\Upsilon'^{(m)} \triangleq \{\mathcal{T}'^{(m,1)}, \text{perm}_{(2,1,3)}(\mathcal{T}'^{(1,m)*}), \ldots, \mathcal{T}'^{(m,M)}, \text{perm}_{(2,1,3)}(\mathcal{T}'^{(M,m)*})\}. \quad (S2)$$

The $2M$ tensors in $\Upsilon'^{(m)}$ can be further grouped into $M$ pairs: $\{\mathcal{T}'^{(m,n)}, \text{perm}_{(2,1,3)}(\mathcal{T}'^{(n,m)*})\}$, with fixed $m$ and varying $n$. We concatenate each pair of tensors along the third mode into a new tensor that admits a PD, as:

$$\mathcal{T}^{(m,n)} \triangleq \begin{cases} \text{cat}(\mathcal{T}'^{(m,n)}, \text{perm}_{(2,1,3)}(\mathcal{T}'^{(n,m)*})) \\ = [\![A^{(m)}, A^{(n)*}, [C'^{(m,n)}, C'^{(n,m)*}]]\!]_R & m < n, \\ \text{cat}(0.5[\mathcal{T}'^{(m,m)} + \text{perm}_{(2,1,3)}(\mathcal{T}'^{(m,m)*})], \\ -0.5\mathrm{i}[\mathcal{T}'^{(m,m)} - \text{perm}_{(2,1,3)}(\mathcal{T}'^{(m,m)*})]) \\ = [\![A^{(m)}, A^{(m)*}, [\text{re}(C'^{(m,m)}), \text{im} C'^{(m,m)}]]\!]_R & m = n, (S3) \\ \text{cat}(\text{perm}_{(2,1,3)}(\mathcal{T}'^{(n,m)*}), \mathcal{T}'^{(m,n)}) \\ = [\![A^{(m)}, A^{(n)*}, [C'^{(n,m)*}, C'^{(m,n)}]]\!]_R & m > n. \end{cases}$$

The new set of tensors constructed by (S3) admits a DC-CPD with conjugated symmetry:

$$\begin{cases} \mathcal{T}^{(m,n)} = [\![A^{(m)}, A^{(n)*}, C^{(m,n)}]\!]_R, \\ \text{where } \mathcal{T}^{(m,n)} = \text{perm}_{(2,1,3)}(\mathcal{T}^{(n,m)*}). \end{cases} \quad (S4)$$

The concatenated tensors $\mathcal{T}^{(m,n)}$ have the advantage over $\mathcal{T}'^{(m,n)}$ that their third dimension is larger, which will make it easier to satisfy the full column rank assumption of the factor matrix in the third mode, as imposed by the derivation in *Section III* of the main manuscript.

We use a numerical validation to show the above-mentioned merit. We generate a set of tensors $\{\mathcal{T}'^{(m,n)} \in \mathbb{C}^{N \times N \times T}\}$ that admit a non-symmetric DC-CPD: $\mathcal{T}'^{(m,n)} = [\![A^{(m)}, A^{(n)*}, C'^{(m,n)}]\!]_R$, where both the real and imaginary parts of each entry of $A^{(m)}, C'^{(m,n)}$ are randomly drawn from a standard normal distribution. We let $N = 4$, $T = 4$, and the double coupled rank $R = 6$. Note that all the three dimensions of each tensor are less than $R$. The proposed algebraic DC-CPD algorithm does not work for $\{\mathcal{T}'^{(m,n)}\}$ as the third factor matrices do not have full column rank. Then we create a set of conjugated symmetric tensors $\{\mathcal{T}^{(m,n)}\}$ by (S3), note that the third factor matrices are of size $8 \times 6$, and thus have full column rank generically. We apply the proposed algebraic DC-CPD algorithm to $\{\mathcal{T}^{(m,n)}\}$, and the algorithm finds the exact solution with mean relative error $\varepsilon = 1.21 \times 10^{-12}$.

## II. Derivation of Equation (15)

Recall that the $N^4 \times R^2$ matrix $\Gamma^{(m,g,h)}$ holds the vectors $\psi(T^{(m,g)}_{(:,:,s)}, T^{(m,h)}_{(:,:,u)})$, with fixed $(m, g, h)$ and varying $s, u$, as its columns:

$$\left(\Gamma^{(m,g,h)}\right)_{:,(s-1)R+u} = \psi\left(T^{(m,g)}_{(:,:,s)}, T^{(m,h)}_{(:,:,u)}\right), \quad (S5)$$

where $\psi(T^{(m,g)}_{(:,:,s)}, T^{(m,h)}_{(:,:,u)}) = \text{vec}(\Phi(T^{(m,g)}_{(:,:,s)}, T^{(m,h)}_{(:,:,u)}))$, and $\Phi(\cdot)$ is the coupled rank-1 detection mapping, defined in (12) as:

$$\left[\Phi(X^{(1)}, X^{(2)})\right]_{i,j,p,q} \triangleq \begin{vmatrix} x^{(1)}_{i,p} & x^{(2)}_{i,q} \\ x^{(1)}_{j,p} & x^{(2)}_{j,q} \end{vmatrix} = x^{(1)}_{i,p} x^{(2)}_{j,q} - x^{(1)}_{j,p} x^{(2)}_{i,q}. \quad (S6)$$

Due to the bilinearity of $\psi$, we have:

$$\Gamma^{(m,g,h)} = \sum_{t,r=1}^{R} \psi(a^{(m)}_t a^{(g)H}_r, a^{(m)}_r a^{(h)H}_r) \cdot (c^{(m,g)}_t \otimes c^{(m,h)}_r)^T. \quad (S7)$$

According to *Theorem 1*, $\psi(a^{(m)}_t a^{(g)H}_r, a^{(m)}_r a^{(h)H}_r) = \mathbf{0}$ when $t = r$ and thus (S7) can be rewritten as:

$$\Gamma^{(m,g,h)} = \sum_{t \neq r} \psi(a^{(m)}_t a^{(g)H}_r, a^{(m)}_r a^{(h)H}_r) \cdot (c^{(m,g)}_t \otimes c^{(m,h)}_r)^T. \quad (S8)$$

We stack the vectors $c^{(m,g)}_t \otimes c^{(m,h)}_r$ into a $R^2 \times (R^2 - R)$ matrix $P^{(m)}_{g,h}$, and the vectors $\psi(a^{(m)}_t a^{(g)H}_r, a^{(m)}_r a^{(h)H}_r)$ into an $N^4 \times (R^2 - R)$ matrix $\Phi^{(m)}_{g,h}$, where the columns are indexed by $(t, r)$, $1 \leq t \neq r \leq R$. Then (S8) can be rewritten as $\Gamma^{(m,g,h)} = \Phi^{(m)}_{g,h} \cdot P^{(m)T}_{g,h}$ (equation (15) in the paper).

## III. DETERMINATION OF GENERIC VALUE OF $R_{\max}$

*Corollary 10* in [1] states that, for an analytic function f($x$) of multiple complex variable $x = [x_1, ..., x_L]^T \in \mathbb{C}^L$, if f($x$) is not identically zero, then the set $V = \{x \in \mathbb{C}^L | f(x) = 0\}$ has Lebesque measure zero.

*Theorem 2* in the main manuscript states that a DC-CPD is unique if the following conditions hold for all $1 \leq g < h \leq M$, $1 \leq m, n \leq M$:

(*i*) Factor matrices $C^{(m,n)}$ have full column rank

(*ii*) Vectors $\psi(a_t^{(m)} a_t^{(g)H}, a_r^{(m)} a_r^{(h)H})$ are linear independent for all $1 \leq t \neq r \leq R$.

Generically, sub-statement (*i*) suggests that the number of rows of matrices $C^{(m,n)}$ is not less than its columns. Hence, the third dimension of the tensors is required to be not less than $R$.

Now we explain the generic version of sub-statement (*ii*). We note that this sub-statement implies that the matrix $\Phi_{g,h}^{(m)}$ should have full column rank. The matrix $\Phi_{g,h}^{(m)}$ is of size $N^4 \times (R^2 - R)$ and holds all the vectors $\psi(a_t^{(m)} a_t^{(g)H}, a_r^{(m)} a_r^{(h)H})$ as its columns for $1 \leq t \neq r \leq R$.

We consider a vector-valued function $\mathbf{f}_{g,h}^{(m)}(x)$ that holds all the $(R^2 - R) \times (R^2 - R)$ minors of $\Phi_{g,h}^{(m)}$. The vector $x$ holds all the entries of $A^{(m)}, m = 1, ..., M$ as variables. By definition, $\mathbf{f}_{g,h}^{(m)}(x) = \mathbf{0}$ if $\Phi_{g,h}^{(m)}$ does not have full column rank.

According to *Corollary 10* in [1], if there exists $x_0$ such that $\mathbf{f}_{g,h}^{(m)}(x_0) \neq \mathbf{0}$, then $\mathbf{f}_{g,h}^{(m)}(x) \neq \mathbf{0}$ "almost everywhere". In other words, if we can find one example such that $\Phi_{g,h}^{(m)}$ has full column rank, then $\Phi_{g,h}^{(m)}$ has full column rank generically.

Therefore, to determine the generic value of $R_{\max}$, we can just randomly generate factor matrices $A^{(1)}, ..., A^{(M)}$ for a given $R$, and check if $\Phi_{g,h}^{(m)}$ has full column rank. $R_{\max}$ is then chosen as the largest $R$ for which the matrices $\Phi_{g,h}^{(m)}$, for all $1 \leq g < h \leq M, 1 \leq m \leq M$, have full column rank.

## IV. DISCUSSION OF THE EFFICIENCY OF ALGEBRAIC DC-CPD

### A. An efficient implementation of algebraic DC-CPD

We note that the main complexity of the algebraic DC-CPD algorithm, as it has been presented in the manuscript, is in the construction of the $N^4 \times R^2$ matrices $\Gamma^{(m,g,h)}$ and the calculation of the basis vectors in their null space. Here we present an efficient way to calculate the basis vectors. In this implementation, we do not explicitly construct $\Gamma^{(m,g,h)}$. Instead, we calculate the $R^2 \times R^2$ matrices $\Omega^{(m,g,h)} \triangleq \Gamma^{(m,g,h)H} \Gamma^{(m,g,h)}$. We note that $\ker(\Omega^{(m,g,h)}) = \ker(\Gamma^{(m,g,h)})$, while $\Omega^{(m,g,h)}$ has smaller size and can be computed in a more efficient way than $\Gamma^{(m,g,h)}$, as will be explained later. Hence, in an efficient implementation we use $\Omega^{(m,g,h)}$ instead of $\Gamma^{(m,g,h)}$.

More precisely, as $\Gamma^{(m,g,h)}$ holds $\psi(T_{(:,:,s)}^{(m,g)}, T_{(:,:,u)}^{(m,h)})$ as its columns, then according to (14) in the manuscript, the entries of $\Omega^{(m,g,h)}$ have the form:

$$[\Omega^{(m,g,h)}]_{(s-1)R+u,(s'-1)R+u'} = \psi^H(T_{(:,:,s)}^{(m,g)}, T_{(:,:,u)}^{(m,h)}) \psi(T_{(:,:,s')}^{(m,g)}, T_{(:,:,u')}^{(m,h)})$$
$$= 2\left(t_s^{(m,g)H} t_{s'}^{(m,h)}\right)\left(t_u^{(m,g)H} t_{u'}^{(m,h)}\right)$$
$$- \sum_{p,q}\left[\left(\tau_{p,s}^{(m,g)H} \tau_{q,s'}^{(m,g)}\right)\left(\tau_{q,u}^{(m,h)H} \tau_{p,u'}^{(m,h)}\right)\right.$$
$$\left. + \left(\tau_{q,s}^{(m,g)H} \tau_{p,s'}^{(m,g)}\right)\left(\tau_{p,u}^{(m,h)H} \tau_{q,u'}^{(m,h)}\right)\right], \quad \text{(S9)}$$

where vectors $t_r^{(m,l)} = \text{vec}(T_{(:,:,r)}^{(m,l)})$ have length $N^2$, and vectors $\tau_{k,r}^{(m,l)} = T_{(k,:,r)}^{(m,l)}$ have length $N$, $l \in \{g, h\}, r \in \{s, s', u, u'\}, k \in \{p, q\}$. We construct two $R \times R$ Hermitian matrices $Z^{(m,l)} \triangleq T_3^{(m,l)H} T_3^{(m,l)}$, where $T_3^{(m,l)}$ is the mode-3 representation of tensor $\mathcal{T}^{(m,l)}$. By definition we have: $Z^{(m,l)}(s, s') = t_s^{(m,l)H} t_{s'}^{(m,l)}$. In addition, we construct $NR \times NR$ Hermitian matrices $Y^{(m,l)} = T_2^{(m,l)} T_2^{(m,l)H}$, where $l \in \{g, h\}$. Index permutation yields $R^2 \times N^2$ matrices $Y_1^{(m,l)}$ and $Y_2^{(m,l)}$:

$$\left[Y_1^{(m,l)}\right]_{(s-1)R+s',(i-1)N+j} = \left[Y^{(m,l)}\right]_{(j-1)R+s',(i-1)R+s},$$
$$\left[Y_2^{(m,l)}\right]_{(s-1)R+s',(j-1)N+i} = \left[Y^{(m,l)}\right]_{(j-1)R+s',(i-1)R+s}. \quad \text{(S10)}$$

Then by definition we know:

$$\left(\Omega^{(m,g,h)}\right)_{(s-1)R+u,(s'-1)R+u'} = 2\left(Z^{(m,g)}\right)_{s,s'}\left(Z^{(m,h)}\right)_{u,u'}$$
$$- \left(Y_1^{(m,g)} Y_2^{(m,h)T} + Y_2^{(m,g)} Y_1^{(m,h)T}\right)_{(s-1)R+s',(u-1)R+u'}. \quad \text{(S11)}$$

In practice, we will precompute $Z^{(m,l)}$, $Y^{(m,l)}$, and construct Hermitian $R^2 \times R^2$ matrices $\Omega^{(m,g,h)}$ by (S10) and (S11).

### B. Analysis of complexity and memory requirements

We first analyze the complexity and memory requirements of the original version of the algebraic DC-CPD algorithm, and then provide results related to its efficient implementation.

In the original algebraic DC-CPD algorithm, we note that the main complexity is in the construction of matrices $\Gamma^{(m,g,h)}$ and the calculation of the basis vectors $w_r^{(m,g,h)}$ in their null space. According to (12) in the manuscript, we have that the complexity of the construction of each column of $\Gamma^{(m,g,h)}$ is $O(2N^4)$ flops. Hence, the overall complexity of the construction of $\Gamma^{(m,g,h)}$ is $O(2N^4 R^2)$ flops. The basis vectors in $\ker(\Gamma^{(m,g,h)})$ are computed as singular vectors. This step has complexity $O(2N^4 R^4 + 11R^6)$ flops. We construct and manipulate $0.5M^2(M-1)$ such matrices $\Gamma^{(m,g,h)}$. The overall complexity is thus $O(M^3 N^4 R^2 + M^3 N^4 R^4 + 5.5M^3 R^6)$ flops. The memory requirements of the algorithm are mainly in the storage of the $0.5M^2(M-1)$ tensors $\mathcal{W}^{(m,g,h)}$ and the matrix $\Gamma^{(m,g,h)}$. Hence, the memory requirements are $O(0.5M^3 R^3 + 0.5M^3 N^4 R^2)$ complex numbers.

In the efficient implementation, the complexity of computing a single Hermitian matrix $Z^{(m,l)}$ is $O(0.5N^2 R^2)$ flops, and the complexity of computing a single Hermitian matrix $Y^{(m,l)}$ is $O(0.5N^3 R^2)$ flops. The additional complexity of computing a single Hermitian matrix $\Omega^{(m,g,h)}$ by (S11) is $O(0.5N^2 R^4 + 0.5R^4)$ flops. If we compute the basis vectors of $\ker(\Omega^{(m,g,h)})$ by modified Gram-Schmidt orthogonalization, then the computational cost is $O(2R^6)$ flops. If we use EVD, then the computational cost is $O(6R^6)$ flops.

For the indices $m, l = 1, ..., M$, $1 \leq g < h \leq M$, the algebraic DC-CPD overall requires the construction of $M^2$ matrices $\mathbf{Z}^{(m,l)}$ of size $R \times R$, $M^2$ matrices $\mathbf{Y}^{(m,l)}$ of size $NR \times NR$, $0.5M^2(M-1)$ matrices $\mathbf{\Omega}^{(m,g,h)}$ of size $R^2 \times R^2$, and the calculation of the $R$ basis vectors of $\ker(\mathbf{\Omega}^{(m,g,h)})$. The overall complexity is $O(0.5M^2N^2R^2 + 0.5M^2N^3R^2 + 0.25M^3R^4 + 0.25M^3N^2R^4 + M^3R^6)$ flops. For the memory requirements we consider the storage of the matrices $\mathbf{Z}^{(m,l)}$, $\mathbf{Y}^{(m,l)}$, $\mathbf{\Omega}^{(m,p,q)}$, and the tensors $\mathcal{W}^{(m,p,q)}$ (17). We need to store $O(M^2R^2 + M^2N^2R^2 + 0.5M^3R^4 + 0.5M^3R^3)$ complex numbers. The above expressions for complexity and memory requirements can be simplified to $O(0.5M^3N^2R^4 + M^3R^6)$ flops and $O(M^2N^2R^2 + 0.5M^3R^4)$ complex numbers, respectively.

## V. RESULTS OF EXPERIMENT B

Besides the plots in the manuscript, we provide more results for experiment B, in Fig.1–Fig. 3. The plots in Fig. 1 report the mean relative error versus SNR, while the plots in Fig. 2 report the mean CPU time versus SNR. The plots concern the following three cases: (a) overdetermined J-BSS: $N = 3$, $R = 3$; (b) slightly underdetermined J-BSS: $N = 3$, $R = 4$; (c) highly underdetermined J-BSS: $N = 4$, $R = 8$. In each case, parameters such as $L$, $T$, $M$ are varied.

The plots in Fig.3 report the mean CPU time versus $M$. The following three settings are considered: (a) an overdetermined setting with $N = 3$, $R = 3$, $L = 250$, $T = 39$, SNR = 10dB; (b) a slightly underdetermined setting with $N = 3$, $R = 4$, $L = 250$, $T = 39$, SNR = 10dB; (c) a highly underdetermined setting with $N = 4$, $R = 8$, $L = 250$, $T = 39$, SNR = 20dB. In all these settings we vary $M$ from 2 to 10.

With more results for different parameters, we aim to: (a) provide a more comprehensive comparison of the performance of the algorithms, (b) provide insights into how the performance of different algorithms depends on the parameters, and (c) provide a general idea on the computational efficiency of the compared algorithms. Please refer to the manuscript for the details of the data generation and the settings for the compared algorithms. For convenience, we put the three plots in Fig. 3 of the manuscript in the first row of Fig.1.

### A. Mean relative error versus SNR

From Fig.1, we first draw a few general conclusions about the relative performance of the compared algorithms. In the overdetermined case, with subfigures (1.1), (1.4), (1.7) of Fig.1, we see that the DC-CPD algorithms perform better than CPD-C and SOBIUM. GOJD has the best performance at high SNR, and poorer performance than DC-CPD at low SNR. This behavior is consistent for different parameters, and the good performance of GOJD at high SNR is mainly because of the prewhitening step used in this algorithm. MCCA always performs poorly, because it does not make use of the temporal structure of the signals. Comparing the four DC-CPD algorithms, we note that the results of DC-CPD-ALG are not much improved by the optimization based algorithms in this relatively easy case of overdetermined J-BSS.

In the underdetermined cases, we consistently observe that DC-CPD algorithms perform better than SOBIUM and CPD-C. The comparison between DC-CPD and SOBIUM and CPD-C, in both overdetermined and underdetermined cases, shows that it is in general advantageous to take more of the coupling structure into account. The difference between DC-CPD and SOBIUM and CPD-C is more pronounced in cases that are more difficult. Comparing the four DC-CPD algorithms, we see that the optimization based DC-CPD algorithms, DC-CPD-ALS, DC-CPD-SDF(NLS), and DC-CPD-SDF(QN) are more accurate than DC-CPD-ALG. Note that the improvement by the optimization based DC-CPD algorithms is more significant in the more challenging underdetermined cases, than in the overdetermined case. The above general observations are consistent with those reported in the manuscript.

Now we examine how the attained accuracy depends on the choice of parameter values. In the overdetermined case, we note that the performance is improved by increasing either the frame length $L$ or the number of frames $T$, as shown in subfigures (1.1), (1.4) and (1.7), but the improvement is not as much as in the underdetermined case. This is because the overdetermined problem is relatively easy and less sensitive to noise and finite sample effects. The compared algorithms perform well in general, even if the frame length is very small (note that $L = 50$ in subfigure (1.1)).

On the other hand, in the highly underdetermined case, with subfigures (1.3), (1.6) and (1.9), the problem itself becomes very sensitive to noise and finite sampling errors. In particular, when $L = 250$, $T = 19$, $M = 3$, as shown in subfigure (1.6), most of the compared algorithms do not generate reasonable results even at high SNR. When $L = 250$, $T = 39$, $M = 3$, as shown in subfigure (1.3), the results of the DC-CPD algorithms are reasonably accurate at high SNR. CPD-C and SOBIUM, on the other hand, do not perform as accurate as DC-CPD at all SNR levels. When $L = 1000$, $T = 79$, $M = 4$, as shown in subfigure (1.9), all the compared algorithms perform well at high SNR. The DC-CPD algorithms even generate very accurate results (mean relative error < 0.1) at a medium level of SNR (e.g. 12dB). However, this requires a large number of samples (note that in this setting the total number of samples equals 40000). The above observations generally illustrate that the highly underdetermined problem is by itself very sensitive to noise and finite sample effects.

In the slightly underdetermined case, we observe that all the compared algorithms are able to generate reasonably accurate results for medium level of SNR, and medium values of $L$ and $T$, as shown in subfigures (1.2), (1.5), and (1.11). Comparing subfigures (1.8) and (1.10), we see that the decrease of $T$ results in a large deterioration of the accuracy. In particular, in subfigure (1.8), all the compared algorithms perform reasonably well at high SNR. In subfigure (1.10), none of the algorithms yields good results. Comparing subfigures (1.2), (1.5) and (1.11), we note that the increase of $L$ improves the performance of all the compared algorithms.

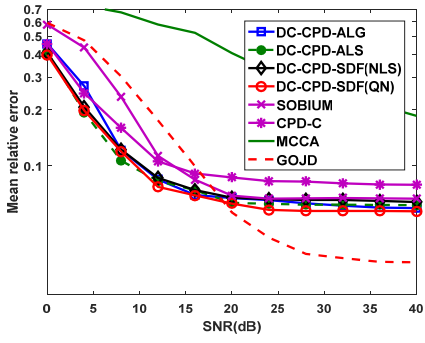
(1.1) $N = 3, R = 3, M = 3, L = 50, T = 39$

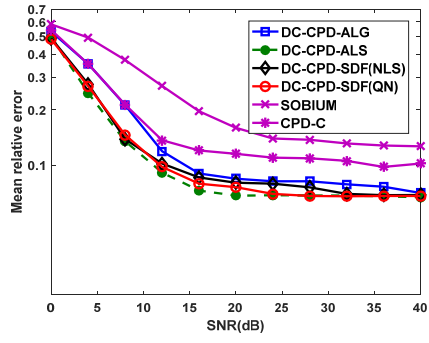
(1.2) $N = 3, R = 4, M = 3, L = 100, T = 39$

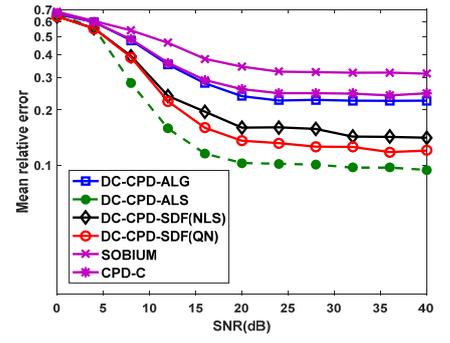
(1.3) $N = 4, R = 8, M = 3, L = 250, T = 39$

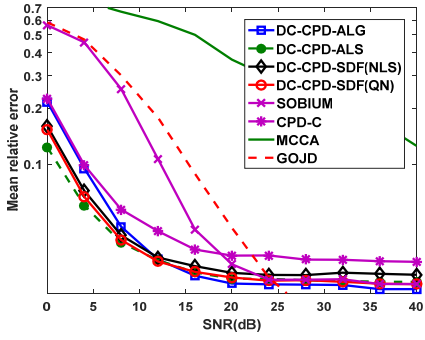
(1.4) $N = 3, R = 3, M = 3, L = 250, T = 39$

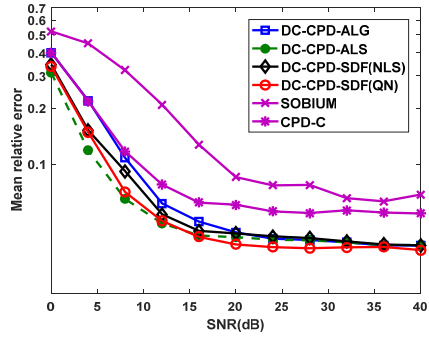
(1.5) $N = 3, R = 4, M = 3, L = 250, T = 39$

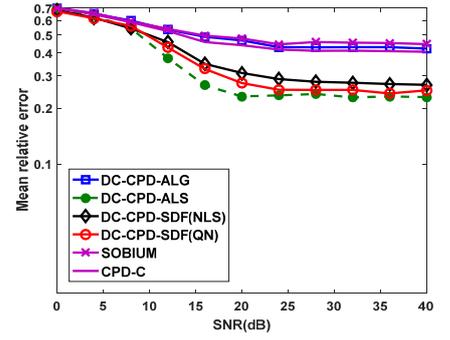
(1.6) $N = 4, R = 8, M = 3, L = 250, T = 19$

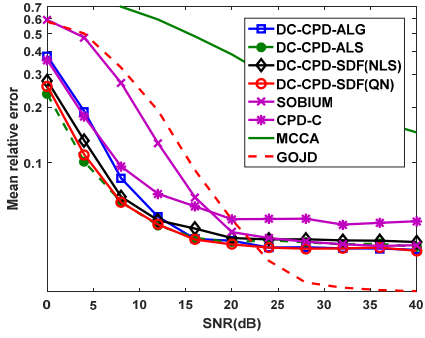
(1.7) $N = 3, R = 3, M = 3, L = 250, T = 19$

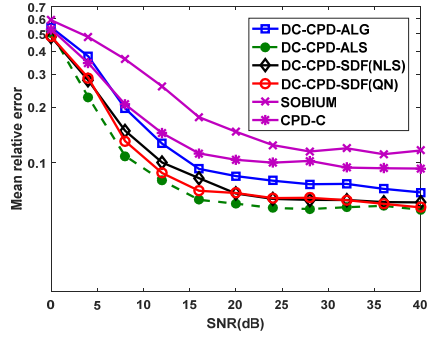
(1.8) $N = 3, R = 4, M = 3, L = 250, T = 19$

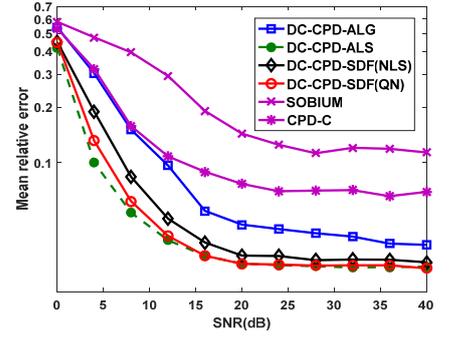
(1.9) $N = 4, R = 8, M = 3, L = 1000, T = 79$

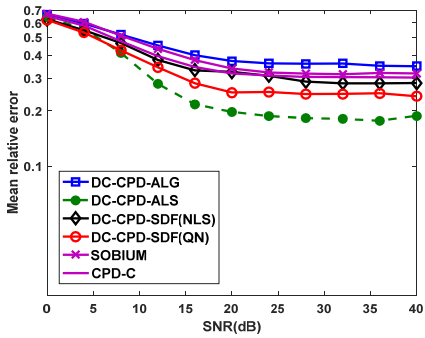
(1.10) $N = 3, R = 4, M = 3, L = 250, T = 7$

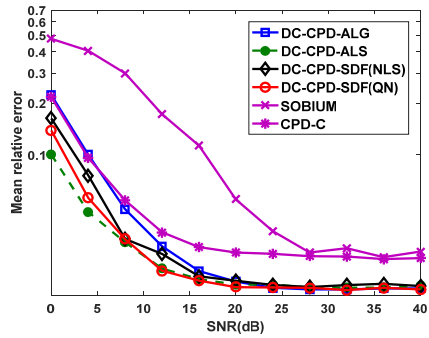
(1.11) $N = 3, R = 4, M = 3, L = 1000, T = 39$

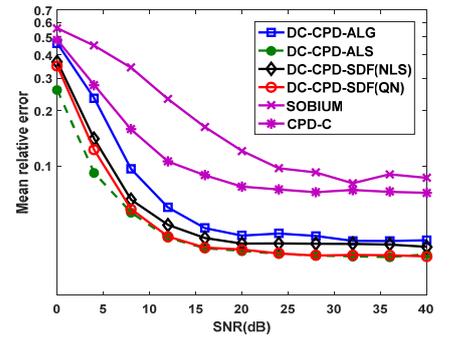
(1.12) $N = 3, R = 4, M = 5, L = 250, T = 39$

Fig. 1. Mean relative error of DC-CPD-ALG, DC-CPD-ALS, DC-CPD-SDF(NLS), DC-CPD-SDF(QN), SOBIUM, and CPD-C vs. SNR in experiment B, in setting (a) MCCA and GOJD are also included in the comparison. The plots illustrate the performance in (a) an overdetermined setting, (b) a slightly underdetermined setting, and (c) a highly underdetermined setting.

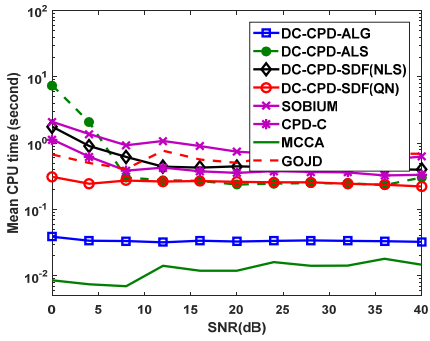
(2.1) $N = 3$, $R = 3$, $M = 3$, $L = 50$, $T = 39$

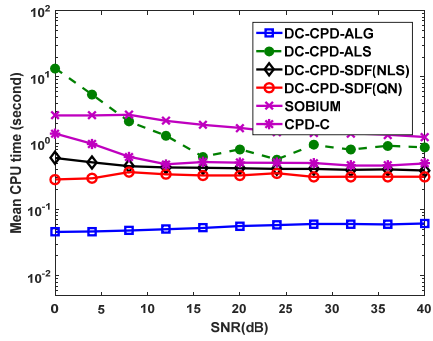
(2.2) $N = 3$, $R = 4$, $M = 3$, $L = 100$, $T = 39$

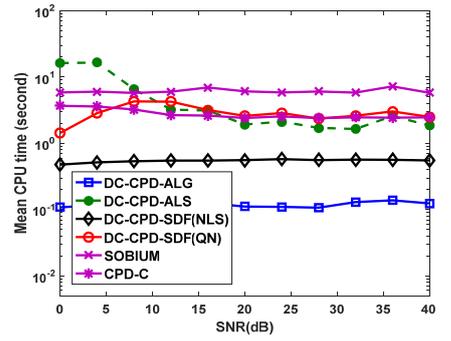
(2.3) $N = 4$, $R = 8$, $M = 3$, $L = 250$, $T = 39$

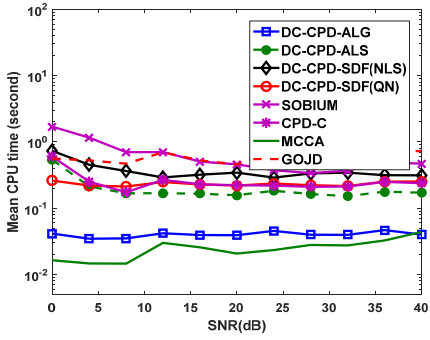
(2.4) $N = 3$, $R = 3$, $M = 3$, $L = 250$, $T = 39$

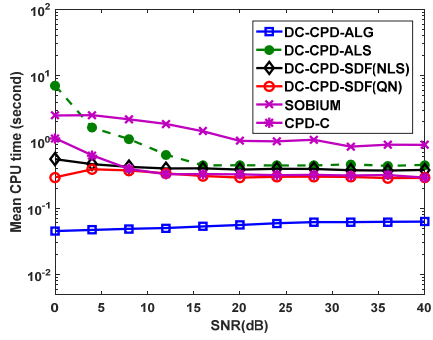
(2.5) $N = 3$, $R = 4$, $M = 3$, $L = 250$, $T = 39$

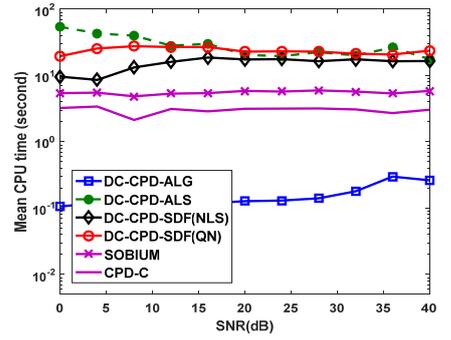
(2.6) $N = 4$, $R = 8$, $M = 3$, $L = 250$, $T = 19$

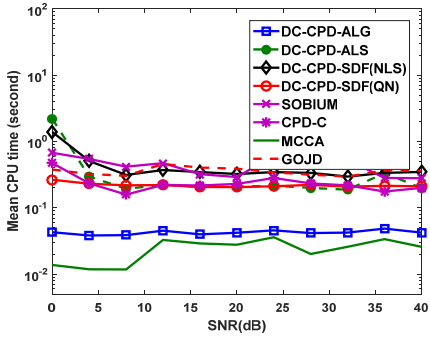
(2.7) $N = 3$, $R = 3$, $M = 3$, $L = 250$, $T = 19$

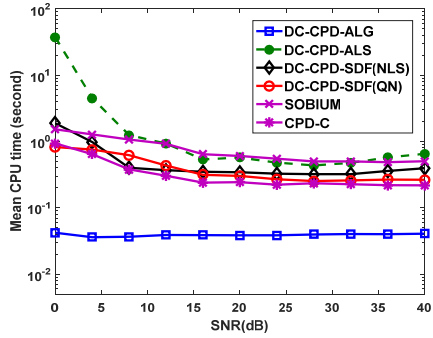
(2.8) $N = 3$, $R = 4$, $M = 3$, $L = 250$, $T = 19$

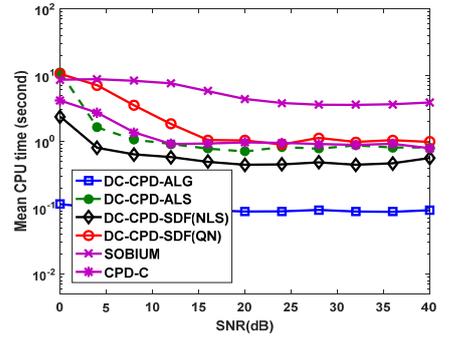
(2.9) $N = 4$, $R = 8$, $M = 3$, $L = 1000$, $T = 79$

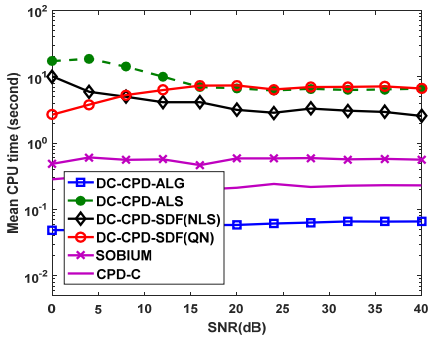
(2.10) $N = 3$, $R = 4$, $M = 3$, $L = 250$, $T = 7$

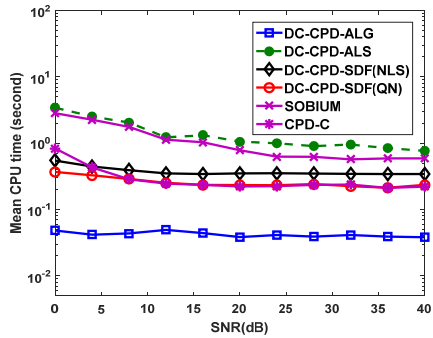
(2.11) $N = 3$, $R = 4$, $M = 3$, $L = 1000$, $T = 39$

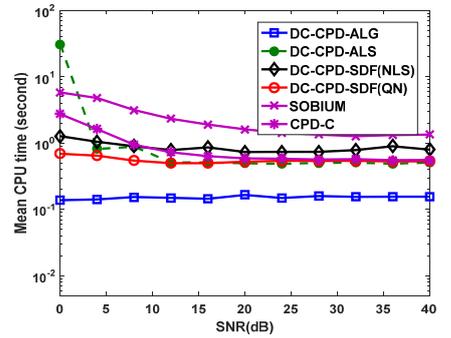
(2.12) $N = 3$, $R = 4$, $M = 5$, $L = 250$, $T = 39$

Fig. 2. Mean CPU time of DC-CPD-ALG, DC-CPD-ALS, DC-CPD-SDF(NLS), DC-CPD-SDF(QN), SOBIUM, and CPD-C vs. SNR in experiment B, in setting (a) MCCA and GOJD are also included in the comparison. The plots illustrate the mean CPU time in (a) an overdetermined setting, (b) a slightly underdetermined setting, and (c) a highly underdetermined setting.

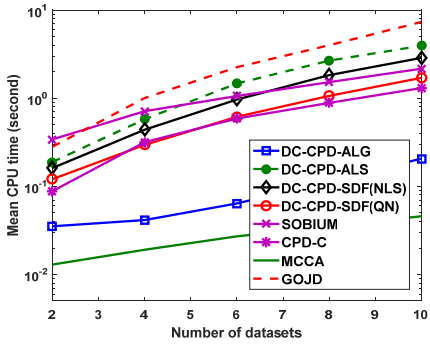 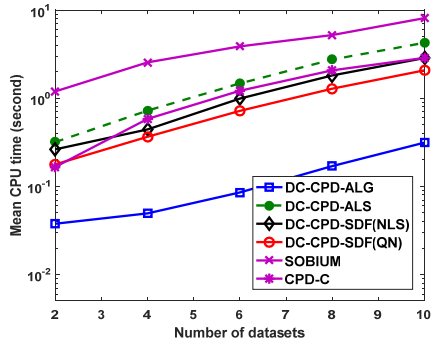 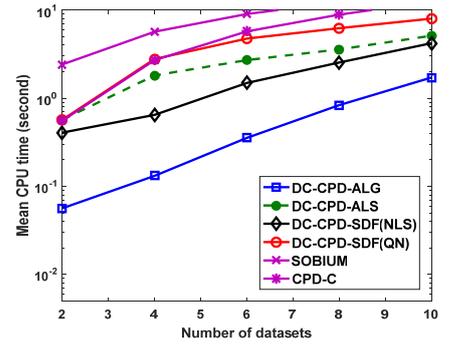

(3.1) $N = 3$, $R = 3$, $L = 250$, $T = 39$, SNR = 10dB   (3.2) $N = 3$, $R = 4$, $L = 250$, $T = 39$, SNR = 10dB   (3.3) $N = 4$, $R = 8$, $L = 250$, $T = 39$, SNR = 20dB

Fig. 3. Mean CPU time of DC-CPD-ALG, DC-CPD-ALS, DC-CPD-SDF(NLS), DC-CPD-SDF(QN), SOBIUM, and CPD-C vs. the number of datasets $M$, in setting (a) MCCA and GOJD are also included in the comparison. The plots illustrate the performance in (a) an overdetermined setting, (b) a slightly underdetermined setting, and (c) a highly underdetermined setting.

### B. Mean CPU time versus SNR and M

In Fig.2, we observe that the mean CPU time of DC-CPD-ALG is less than that of other compared algorithms. The only exceptions are subfigure (2.1), (2.4), and (2.7) in the overdetermined case, where DC-CPD-ALG is slightly slower than MCCA. We also observe that the mean CPU time of DC-CPD-ALG does not vary much with $L$, $T$, and SNR. This is because the complexity of an algebraic algorithm only depends on the size of the problem, e.g. $N$, $R$, $M$, as suggested in our complexity analysis in *Subsection V.D* of the manuscript. Moreover, we observe that the mean CPU time of DC-CPD-SDF(NLS) is less than that of DC-CPD-ALS in the underdetermined cases. This is thanks to the low per-iteration cost and quadratic convergence of NLS iterations, and is consistent with other observations for problems that are somewhat challenging [2].

In Fig. 3, we observe that the mean CPU time of all the compared algorithms increases when $M$ is larger. In all the three settings, DC-CPD-ALG is much faster than DC-CPD-ALS, DC-CPD-SDF (NLS), and DC-CPD-SDF(QN). In the overdetermined setting, we include MCCA and GOJD in the comparison, DC-CPD-ALG is faster than GOJD, while slightly slower than MCCA. However, as shown in Fig.1, DC-CPD-ALG is much more accurate than MCCA for the considered J-BSS problem.